\documentclass[10pt,twocolumn,letterpaper]{article}

\usepackage[
]{cvpr}    

\definecolor{cvprblue}{rgb}{0.21,0.49,0.74}
\definecolor{bluearrow}{rgb}{0.4, 0.6, 0.9} 
\definecolor{brightpink}{rgb}{1.0, 0.0, 0.5}
\definecolor{orangecolor}{rgb}{1.0, 0.5, 0.0} 
\definecolor{brightgreen}{rgb}{0.0, 0.8, 0.2}

\usepackage{amsmath}
\usepackage{graphicx}
\usepackage[normalem]{ulem}  

\usepackage{balance}
\usepackage{multirow} 

\usepackage{adjustbox}

\usepackage{float}
\usepackage{booktabs}
\usepackage{array}
\usepackage{caption}
\usepackage{wrapfig}
\usepackage{lipsum} 
\usepackage{microtype}
\usepackage[breaklinks,colorlinks,allcolors=cvprblue]{hyperref}
\makeatletter
\def\@LN@col#1{}
\makeatother
\def\paperID{13771} 
\def\confName{CVPR}
\def\confYear{2025}

\title{DiffMVR: Diffusion-based Automated Multi-Guidance Video Restoration}

\author{Zheyan Zhang\\
Northwestern University\\
Department of Industrial Engineering and Management Sciences\\
{\tt\small ZheyanZhang2027@u.northwestern.edu}
\and
Diego Klabjan\\
Northwestern University\\
Department of Industrial Engineering and Management Sciences\\
{\tt\small d-klabjan@northwestern.edu}
\and
Renee CB Manworren\\
University of Texas at Arlington\\
College of Nursing and Health Innovation\\
{\tt\small Renee.Manworren@uta.edu}
}

\begin{document}
\maketitle
\begin{abstract}
In this work, we address a challenge in video inpainting: reconstructing occluded regions in dynamic, real-world scenarios. Motivated by the need for continuous human motion monitoring in healthcare settings, where facial features are frequently obscured, we propose a diffusion-based video-level inpainting model, DiffMVR. Our approach introduces a dynamic dual-guided image prompting system, leveraging adaptive reference frames to guide the inpainting process. This enables the model to capture both fine-grained details and smooth transitions between video frames, offering precise control over inpainting direction and significantly improving restoration accuracy in challenging, dynamic environments. DiffMVR represents a significant advancement in the field of diffusion-based inpainting, with practical implications for real-time applications in various dynamic settings.
\end{abstract}

\section{Introduction}
\label{sec:intro}
The rise of diffusion models has revolutionized computer vision, driving advances in image editing, super-resolution, object removal, and restoration. Diffusion-based inpainting, in particular, has seen wide application across fields. In medical imaging, these models aid in anomaly detection by restoring diseased regions to a healthy state for comparative analysis ~\cite{wolleb2022diffusionmodelsmedicalanomaly}. In autonomous driving, they reconstruct occluded information like road signs ~\cite{liu2024ddmlagdiffusionbaseddecisionmaking}, while in advertising, they create immersive VR scenes for product promotion ~\cite{Asija20243DPI}. Similarly, for privacy preservation, they assist in the removal of sensitive visual information, such as faces or personal details in shared data. In healthcare, real-time facial action monitoring is crucial for accurate pain assessments ~\cite{HERR2024}. Despite this growing demand for inpainting, most state-of-the-art models are frame-based, achieving high-quality restoration in static images but failing to capture the dynamic essentials for continuous video tasks. As a result, there is a clear need for video-level diffusion models capable of addressing these limitations.

While image-based inpainting methods excel at restoring missing regions in static frames, these models are inherently limited to static 2D content. Foundational models like Denoising Diffusion Probabilistic Models (DDPM)~\cite{NEURIPS2020_4c5bcfec} and variants such as Denoising Diffusion Implicit Models (DDIM)~\cite{song2022denoisingdiffusionimplicitmodels}, introduce robust frameworks to iteratively denoise and restore image content with impressive quality. Other approaches, like Vector Quantized Variational AutoEncoder~\cite{razavi2019generatingdiversehighfidelityimages}, enable high-resolution generation by learning quantized embeddings, yet are restricted to single-frame synthesis. Partial convolution inpainting~\cite{liu2018imageinpaintingirregularholes} addresses irregular masks by conditioning convolutional filters on valid pixels alone, reducing artifacts in static inpainting tasks. Although these models achieve realistic object replacement and seamless integration within static frames, they lack the temporal coherence required for video-level tasks. This creates an opportunity to extend diffusion-based inpainting methods to handle sequential frames in videos, requiring new strategies that address both spatial and temporal aspects.

Building on image-based models, recent advancements in video-level inpainting have targeted the reconstruction of missing or occluded regions in sequences-challenges that 2D image inpainting alone cannot fully overcome. Advances in deep learning have driven substantial progress in video inpainting. For example, Ouyang \etal~\cite{ouyang2021internalvideoinpaintingimplicit} utilize the convolutional neural networks for video inpainting, preserving high-frequency details. Further advancements, such as the First Frame Filling video inpainting model~\cite{lee2024videodiffusionmodelsstrong}, leverage diffusion models to achieve accurate object removal, even with large masks. More recent models, including the Any-length video inpainting model~\cite{zhang2024avidanylengthvideoinpainting} and MotionAura ~\cite{susladkar2024motionaurageneratinghighqualitymotion}, introduce diffusion-based video inpainting frameworks that support various video lengths and inpainting tasks. However, existing models often struggle to accurately reconstruct subtle human motions across frames.

Furthermore, most existing methods primarily concentrate on removing objects from videos, rather than replacing them, especially with precise, detailed replacements. Moreover, to the best of our knowledge, no prior work focuses on dual-image-guided video inpainting. This approach is crucial for tasks such as restoring facial movements, where both removal and accurate restoration are essential.

In this paper, we present DiffMVR, a novel diffusion-based framework for dynamic, pairwise image-guided video inpainting. Our approach introduces two adaptive guiding images to steer inpainting precisely across detailed and complex video sequences. 

To enhance the inpainting process for object obscuration removal and restoration, DiffMVR automatically generates two guidance images for each video frame with occlusions: a symmetric image and a past unobstructed frame. The symmetric image, created by mirroring the visible half of the frame along an axis of symmetry, provides structural guidance. The past unobstructed frame is identified by a fine-tuned YOLOv8 model, which searches for the most recent fully visible object in previous frames. These frames are processed by separate CLIP models to extract key-value pairs. The current masked frame is then encoded into a latent space by a VAE, where random noise is added to serve as a query. This query interacts with the key-value pairs from the guidance images to generate dual attention scores that are weighted and fused. The U-Net uses this combined attention, alongside standard diffusion inputs, to iteratively denoise and recover the clean latent vector, which the VAE finally decodes. By merging the effects of these two guidance images within the U-Net's architecture, DiffMVR effectively integrates spatial details and temporal dynamics, a key innovation that sets our technique apart from prior video inpainting methods.

Building on this foundation, we introduce a motion loss term to further enhance temporal consistency across consecutive frames during the denoising process. This non-separable frame loss function ensures continuity between adjacent noisy representations of video frames, tightly linking each frame to its predecessors and successors, thereby creating a continuous and coherent video stream. Unlike traditional image-level denoising methods that treat each frame in isolation, this approach ensures a unified sequence with seamlessly integrated spatial and temporal elements. This not only preserves the fluidity and natural progression of actions but also faithfully reconstructs the video to authentically represent the original scene’s dynamics and aesthetics.

The contributions of our approach are fivefold. (1) It proficiently captures intricate motions, such as fine facial details and nuanced dynamic content, addressing long-standing challenges in video inpainting. (2) Instead of using a static prompt, our framework dynamically adjusts two guiding images per frame by automatically selecting an unobstructed reference from previous frames and generating a symmetric version of the current frame. This real-time, adaptive strategy substantially enhances inpainting accuracy and temporal coherence, even in demanding scenarios. (3) We propose a novel architecture that combines structural and temporal guidance based on their relevance. Our framework processes the symmetric frame and past unobstructed frame in parallel attention pipelines, intelligently fusing their attention scores within the U-Net structure to provide comprehensive spatio-temporal guidance. (4) We introduce a hybrid loss function nested within the diffusion process, utilizing U-Net's unique layered structure to synergistically merge denoising and motion-consistency terms. This innovation allows for effective feature extraction from both present and neighboring frames, enhancing the U-Net’s ability to dynamically synthesize missing content. Our method harnesses the power of diffusion for progressive frame restoration and optimizes the interaction between structural and temporal data, setting a new standard for precision in video inpainting. (5) Through quantitative and qualitative comparisons, we demonstrate that DiffMVR consistently outperforms state-of-the-art inpainting models. This work paves the way for more robust and reliable AI-driven video inpainting, improving decision-making in real-world scenarios.

Upon acceptance of the paper, we will open source the code, but not the IRB-approved dataset.

\section{Related Work}
\label{sec:formatting}
Images are a crucial medium for information dissemination, but they are often susceptible to noise, damage, and interference, which can impede data analysis and knowledge extraction. To restore damaged images and design images according to human intent, various image inpainting approaches have emerged in recent years. 

Generative Adversarial Networks (GANs)~\cite{GAN} represent a breakthrough in image editing, offering the ability to generate high-quality, realistic images and supporting unsupervised learning. However, despite their strengths, GANs face significant challenges, including training instability and high demand for large datasets, which limits their utility in domains such as historical image restoration, where data is scarce. Following the rise of GANs, a patch-based method ~\cite{zhou2016patchbasedtexturesynthesisimage} was developed to synthesize textures from undamaged regions. While effective for simpler tasks like background subtraction, patch-based models struggle with larger missing regions and maintaining global coherence in complex scenes.

To mitigate some of these limitations, Pathak \etal~\cite{pathak2016contextencodersfeaturelearning} introduce the encoder-decoder structure, marking a significant step forward by offering a more stable approach to fill in missing regions. More recently, diffusion-based models such as the DDPM~\cite{NEURIPS2020_4c5bcfec} and the DDIM~\cite{song2022denoisingdiffusionimplicitmodels} address the shortcomings of GANs. These models resolve issues like mode collapse and offer greater flexibility in handling complex distributions. By employing iterative noise addition and removal processes, diffusion models can generate high-quality inpainted images with enhanced stability and consistency.

Initially, diffusion models face challenges in learning effectively from unmasked surrounding pixels~\cite{saharia2022paletteimagetoimagediffusionmodels}. To tackle this constraint, text-guided models~\cite{Rombach_2022_CVPR} emerge, which allow for more precise user-guided edits by incorporating prompts. These advancements have expanded the capabilities of image inpainting models, enabling them to handle more complex tasks, such as producing high-quality results in difficult scenarios and allowing for exact, user-directed modifications.

As the field of image inpainting has matured, extending these techniques to videos has become a natural progression, driven by applications in fields such as remote sensing, medical diagnosis, and traffic video recovery. Video inpainting introduces unique challenges, including maintaining temporal coherence across frames and handling motion complexities. Early methods, such as those developed by Li \etal~\cite{li2022endtoendframeworkflowguidedvideo}, employ flow-based techniques and deformable convolution to propagate features and enhance temporal consistency in inpainted video sequences. However, these models are often bounded by their reliance on intermediate flow estimation steps, which can introduce errors that propagate through frames. DNN-based inpaint models, such as Copy-and-paste network~\cite{lee2019copyandpastenetworksdeepvideo} and the context-aggregated network~\cite{li2020shorttermlongtermcontextaggregation}, address context restoration through a copy-and-paste approach, aggregating reference frames effectively. More recently, a diffusion-based video inpainting model, AVID ~\cite{zhang2024avidanylengthvideoinpainting}, focuses on object removal guided by consistent text prompts. While AVID and other diffusion-based models excel at eliminating unwanted objects and imperfections, they encounter challenges in recapturing fine details and seamlessly blending inpainted areas with surrounding regions, particularly in highly dynamic videos.

\begin{figure}[H]
\centering
\includegraphics[width=\columnwidth]{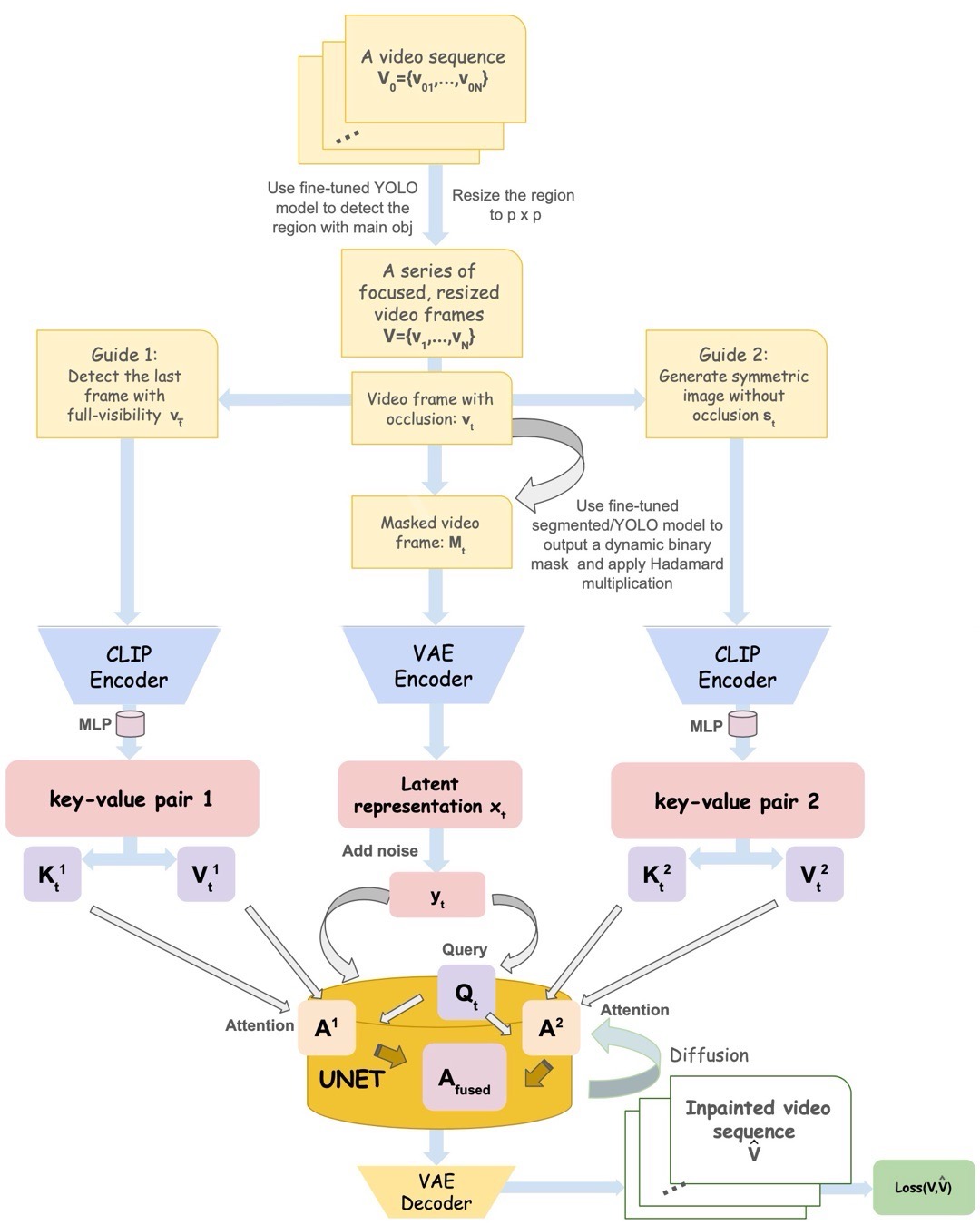}  
\caption{DiffMVR Model Pipeline.}
\label{fig:pip}
\end{figure}

Our proposed methodology tackles these combined challenges through an innovative inpainting pipeline that not only preserves temporal consistency but also thrives in depicting transient features and maintaining structural realism in dynamic areas, such as micro-facial expressions. Utilizing a real-time adaptive guidance framework, our approach dynamically selects and refines guidance images throughout the video sequence, allowing for precise restoration of fine-grained details while preserving both temporal coherence and structural integrity. This evolving dual-guidance design marks a substantial advancement over existing models, delivering more realistic and seamless inpainting even in complex, rapidly changing video scenarios.

\section{Methods}
\label{sec:method}
\subsection{Model Pipeline}
In this section, we establish an automated, multi-image-guided, video-level diffusion-based inpainting pipeline, specifically designed for dynamic video restoration. As illustrated in Figure \ref{fig:pip}, the pipeline consists of four interconnected modules.

The first module $Mod_1$, \textbf{Video Preprocessing}, detects and isolates the primary object in each frame using a fine-tuned YOLO model, ensuring that the inpainting process focuses accurately on regions of interest. This module prepares the input frames by resizing and aligning them for consistent processing. Additionally, we employ a fine-tuned YOLOv8-based model to detect bounding boxes or a segmentation model to identify irregular-shaped occlusions within the object.

The second module, $Mod_2$, \textbf{Visual Encoding}, independently encodes both the frame to be inpainted and its guidance images. The original video frame is processed through a VAE Encoder, which introduces noise to produce a latent representation as input for the diffusion process. Simultaneously, each guidance image, providing structural and temporal cues, is encoded by a CLIP Encoder to generate key-value pairs that facilitate subsequent attention mechanisms.

The third module, $Mod_3$, \textbf{Denoising with Fused Attention}, leverages spatial and temporal cues to guide the U-Net-based denoising process within the diffusion framework. By conditioning on the fused guidance information, this module enhances detail and continuity across frames, improving the quality and consistency of the inpainted video output.

Finally, the fourth module, $Mod_4$, \textbf{Decoding and Restoration}, decodes the fully denoised frame representation back into pixel space using a VAE Decoder, producing the final inpainted frame. Each reconstructed frame is sequentially reassembled into the full video, yielding a temporally consistent inpainted video.

\subsection{Problem Setting}
\label{subsec:set}
We define the input video sequence as \( V_0 = \{v_{0t}\}_{t=1}^N \), which is decomposed into sequential frames. Each frame \( v_{0t} \) undergoes processing to isolate the main object of interest, detected using a fine-tuned YOLOv8 model. The detected object in each frame is subsequently cropped and resized to a uniform resolution of \( p \times p \), producing a refined video sequence \( V = \{v_t\}_{t=1}^N \).

For inpainting facilitation, two guidance images are automatically generated for each frame \( v_t \) where occlusion is present: a symmetric image \( s_t \) and a past unobstructed frame \( v_{\bar{t}} \). The symmetric image \( s_t \) is crafted by mirroring the unoccluded portion of \( v_t \) along an axis of symmetry, defined using Mediapipe for object landmark detection to precisely determine the symmetry line.

The past unobstructed frame \( v_{\bar{t}} \) is sourced through a fine-tuned YOLOv8 model that scans previous frames in \( V \) for the most recently visible object, providing essential temporal guidance.

Additionally, we generate a binary mask \( m_{t,i} \) for each frame \( v_t \), where:
\[
m_{t,i} = \begin{cases}
1, & \text{if pixel } i \text{ is part of the occlusion.} \\
0, & \text{otherwise.}
\end{cases}
\]

In our model, we employ two different mask-generation techniques tailored for continuous video frames. The first mask generation model is based on a YOLOv8 structure and produces bounding box masks. The second model adapts a segmentation-based approach ~\cite{camporese2021HandsSeg2} and provides irregularly contoured masks. These are parts of preprocessing in $Mod_1$. We train and test the pipeline on both types, and both results are presented in Section \ref{sec:exp}.

With the binary masks \( m_{t,i} \) available, we then construct masked video frames as follows:
\[
M_{t} = m_{t,i} \odot v_{t,i}, \forall i \in \text{pixels}, t \in \{2, \dots, N\},
\]
where \(\odot\) denotes the Hadamard product, preserving only the regions indicated by the mask in each frame \( v_t \).

At the end of $Mod_1$, the processed frames \( M_t \), \( s_t \), and \( v_{\bar{t}} \), for \( t \in \{2, \dots, N\} \), \( \bar{t} \in \{1, \dots, {N-1}\} \), are passed to the next module, which encodes spatial and temporal cues into compact representations.

We leverage both the VAE encoder and pre-trained CLIP image embeddings~\cite{radford2021learningtransferablevisualmodels} to extract features for our inpainting pipeline. The masked video frame \( M_t \) is processed by the VAE encoder, transforming it into a spatial latent map \( x_t \). Gaussian noise is then added to this map, producing a noisy latent \( y_t \) as preparation for iterative denoising within the U-Net. 

Simultaneously, the guidance images, namely the symmetric reference \( \{s_t\}_{t=2}^N \) and past unobstructed frames \( \{v_{\bar{t}}\}_{\bar{t}=1}^{N-1} \), are encoded individually using the CLIP encoders. Each guidance image is mapped from its original space to a \( p \)-dimensional feature vector, denoted as \( z_{s_t} \) and \( z_{v_{\bar{t}}} \). 

To ensure compatibility with the dimensions required for the diffusion module, each guidance embedding \( z_{s_t} \) and \( z_{v_{\bar{t}}} \) is passed through a multi-layer perceptron (MLP), \( f_{\text{mlp}}: \mathbb{R}^{p} \rightarrow \mathbb{R}^{p^{\prime}} \), which expands it to a \( p^{\prime} \)-dimensional embedding:
\[
\tilde{z}_{s_t} = f_{\text{mlp}}(z_{s_t}), \quad \tilde{z}_{v_{\bar{t}}} = f_{\text{mlp}}(z_{v_{\bar{t}}}).
\]
\setlength{\abovedisplayskip}{2pt}
\setlength{\belowdisplayskip}{2pt}
The expanded embeddings generate key-value pairs \( (K_t^1, V_t^1) \) and \( (K_t^2, V_t^2) \) from each guidance image independently. These pairs contain spatial and temporal cues, which are then incorporated into the U-Net’s denoising layers through cross-attention. 

In $Mod_3$, at each U-Net layer, a query \( Q_t \) derived from the noisy latent \( y_t \) is used to compute attention scores \( A^1_t \) and \( A^2_t \), representing the relevance of each guidance source:
\[
\hspace{-2pt} \hbox{$ A^1_t = \text{softmax} \left( \frac{Q_t K_t^{1\top}}{\sqrt{D}} \right) V_t^1, \quad A^2_t = \text{softmax} \left( \frac{Q_t K_t^{2\top}}{\sqrt{D}} \right) V_t^2. $}
\]

The final fused attention score \( A_{\text{fused}} \) combines \( A^1_t \) and \( A^2_t \) using weighted coefficients:
\[
A_{\text{fused}}^t = \alpha_1 \cdot A^1_t + \alpha_2 \cdot A^2_t.
\label{eq:fused_embedding}
\]

We employ the dynamically computed \( A_{\text{fused}} \) score at each U-Net denoising layer, guiding the restoration process with high-level structural and temporal context. This innovation has proven its ability to overcome the continuity challenges in video inpainting.

During forward diffusion, noise is incrementally added to \( y_t \), yielding
\[
y_{t,T} = \sqrt{\bar{\alpha}_T} y_{t} + \sqrt{1 - \bar{\alpha}_T} \, \epsilon,
\label{eq:forward_diffusion}
\]
where \( \epsilon \sim \mathcal{N}(0, I) \) represents Gaussian noise, and \( \bar{\alpha}_T = \prod_{i=1}^T \alpha_i \) is the cumulative scaling factor for the noise component for $T=1,2,...$.

The U-Net’s goal is to predict and remove the added noise at each timestep \( T \). The diffusion loss is defined as:
\begin{equation}
\mathcal{L}_{\text{diff}} = \mathbb{E}_{\epsilon \sim \mathcal{N}(0, I), T} \left[ \left\| \epsilon - \epsilon_\theta \left( y_{t,T}, T, A_{\text{fused}}^t \right) \right\|_2^2 \right],
\label{eq:diffloss}
\end{equation}
where \( \epsilon_\theta \left( y_{t,T}, T, A_{\text{fused}}^t \right) \) represents the U-Net’s prediction of the noise component conditioned on the input and fused attention at timestep \( T \).

In the reverse diffusion process, the U-Net iteratively refines \( y_{t,T} \) at each timestep \( T \), aiming to reconstruct \( y_{t,T-1} \):
\[
y_{t,T-1} = \frac{1}{\sqrt{\alpha_T}} \left( y_{t,T} - \frac{1 - \alpha_T}{\sqrt{1 - \bar{\alpha}_T}} \epsilon_\theta \left( y_{t,T}, T, A_{\text{fused}}^t \right) \right) + \sigma_T z,
\label{eq:reverse_diffusion}
\]
where \( z \sim \mathcal{N}(0, I) \) and \( \sigma_T \) represents a noise scale factor at timestep \( T \), adjusting the variance of the noise added back during the reverse diffusion step.

Upon completing the reverse diffusion process, the final denoised latent representation \( \hat{y}_t \) is passed through the VAE decoder to reconstruct the inpainted frame:
\[
\hat{v}_t = D(\hat{y}_t).
\]
These reconstructed frames \( \{\hat{v}_t\}_{t=1}^N \) are then sequentially reassembled to form the final inpainted video sequence \( \hat{V} = \{\hat{v}_t\}_{t=1}^N \), ensuring temporal coherence and spatial fidelity throughout the sequence.

\subsection{Loss Function}
To achieve precise spatial inpainting while maintaining temporal coherence across video frames, we propose a combined loss function. This function is comprised of two components: the denoising loss, which focuses on spatial reconstruction, and the motion-consistency loss, which enforces smooth temporal transitions between frames in video sequences. The combined loss function is defined as
\setlength{\abovedisplayskip}{2pt}
\setlength{\belowdisplayskip}{2pt}
\begin{equation}
\mathcal{L}_{\text{}} = \mathcal{L}_{\text{diff}} + \lambda \cdot \mathcal{L}_{\text{motion}},
\label{eq:total_loss}
\end{equation}
where \( \lambda \) is a weighting factor that balances the impact of temporal coherence against spatial accuracy. 

\noindent
\textbf{Denoising Loss:} The denoising term operates within the diffusion framework, as described in Section \ref{subsec:set}. The primary goal is to restore each masked frame \( M_t \) by iteratively removing noise at each timestep \( T \), which is explicitly shown in ~\eqref{eq:diffloss}.

\noindent
\textbf{Motion-Consistency Loss:} To encourage temporal coherence across consecutive frames during the denoising process, we bring forward a motion-consistency loss term. At each timestep \( T \) of the diffusion process, this loss measures the temporal consistency between adjacent noisy representations of video frames as follows
\begin{equation}
\mathcal{L}_{\text{motion}} = \frac{2}{N} \sum_{t=1}^{N-1} \| y_{t,T} - y_{t-1,T} \|_2^2,
\end{equation}
where \( y_{t,T} \) represents frame \( t \) at diffusion timestep \( T \), and \( N \) is the total number of frames in the current video. This loss term encourages the model to maintain consistent visual features and smooth transitions between consecutive frames while they are being denoised, thereby preventing temporal artifacts that might arise from processing each frame independently.

The motion-consistency loss works in conjunction with the denoising loss throughout the diffusion process, guaranteeing that the final video output exhibits both high-quality spatial reconstruction and smooth temporal dynamics. 

\section{Experiments}
\label{sec:exp}
\subsection{Implementation Details}
Motivated by the obstructions of facial features in babies, which cause inaccurate decision-making based on the monitoring results in healthcare settings, we train and test our framework based on an IRB-approved dataset specifically designed for infant motion monitoring. This Baby dataset contains $120$ videos, each around $20$ seconds in duration, featuring $49$ infants from various ethnic backgrounds. The videos capture a variety of infant statuses including movement, rest, friction, and pain. The dataset also includes $4,101$ images of the same distribution, photographed under diverse lighting conditions, which are used in fine-tuning the YOLO-based detection models. 

Both video frames and images are preprocessed, first centered on the facial region and then resized to $512 \times 512$ pixels. This operation is performed by a fine-tuned YOLOv8-based model, which demonstrates a $100.0\%$ accuracy in detecting the main object, in our case the infant's face.

Our model builds upon the architecture of stable-diffusion-v1-5 checkpoint~\cite{Rombach_2022_CVPR}, with modifications to accommodate the structure of DiffMVR. For each video, we extract frames with $20$ fps, and preprocess the frames following the designation we described in $Mod_1$. Then we choose the ratio of data splitting to be $70\%$ training, $10\%$ validation, and $20\%$ testing.

To evaluate our model's robustness across different occlusion scenarios, we implement two distinct masking approaches. The first uses a fine-tuned YOLOv8n model~\cite{Terven_2023}, trained on annotated images from the Baby dataset, which achieves $98.0\%$ detection accuracy and an average IoU of $0.979$, generating rectangular masks for occlusions. The second method leverages a fine-tuned custom segmentation model~\cite{camporese2021HandsSeg2} trained on $400$ rigorously labeled images, reaching $97.5\%$ accuracy and producing irregular-shaped masks with an average IoU of $0.930$, better mimicking real-world occlusions.

\subsection{Baseline Models and Comparative Analysis}
\label{subsec:baseline}
We evaluate our approach against both image and video inpainting state-of-the-art methods. 

For image-level comparisons, we benchmark against LaMa~\cite{suvorov2021resolutionrobustlargemaskinpainting} for image-guided inpainting, and both original and fine-tuned versions of Stabilityai~\cite{Rombach_2022_CVPR} and Runwayml (an open-source implementation that has been recently removed) models for text-guided inpainting. The fine-tuned variants (\textit{Tuned-stabilityai} and \textit{Tuned-runwayml}) are specifically adapted to our IRB-approved infant dataset. In detail,
\begin{itemize}
    \item \textit{Tuned-stabilityai}: Fine-tuned from a general text-to-image stable diffusion model (Stabilityai) on $10$ static images from $10$ different infants from the Baby dataset over $80$ epochs using Dreambooth.
    \item \textit{Tuned-runwayml}\label{model:runwayml_tune}: Fine-tuned from a text-to-image stable diffusion model (Runwayml) on $3,091$ frames from $40$ infants from the Baby dataset over $250$ epochs.

\end{itemize}

For video-level evaluation, we compare DiffMVR against two advanced video inpainting models. The first is the End-to-End Flow-Guided Video Inpainting (FGVI)~\cite{li2022endtoendframeworkflowguidedvideo}, which leverages flow information for seamless video inpainting. The second model is a deep learning-based approach designed for efficient video restoration (PVI) by Zhou \etal~\cite{zhou2023propainterimprovingpropagationtransformer}. Additionally, to establish a baseline, we independently inpaint each frame using the image-based models Stabilityai, Runwayml and their variants, and LaMa. We then reassemble the frames into videos.

\subsection{Evaluation Metrics}
We evaluate all models both qualitatively and quantitatively, focusing on both the independent images and continuous video frames. To demonstrate the robustness of our pipeline in capturing smooth transitions and restoring intricate details, we choose the following metrics: FID~\cite{ganstrained}, SSIM~\cite{ssimzhou}, TC~\cite{lai2018learningblindvideotemporal}, and FVD~\cite{unterthiner2019accurategenerativemodelsvideo}. These metrics allow us to perform an all-rounded evaluation from three dimensions: structural similarity, the reality of restoration, and temporal coherence, for both frame-level and video-level comparisons. For details on the definition and usage of the metrics, please refer to the Appendix.

\subsection{Quantitative Results}
\subsubsection{Frame-level}

We leverage the $4,101$ images in the Baby dataset for the calculation of SSIM and FID scores. Additionally, we use $120$ videos, each sampled at a frame extraction rate of $20$ frames per second, for the calculation of the TC score. 

To further demonstrate our model's general ability to remove occlusions and restore intricate object details, we introduce the \textbf{HandOverFace (HOF)} dataset as an additional test set. This dataset comprises of $302$ images featuring various hand-over-face scenarios from a different distribution. Collected from publicly available sources, the HOF dataset represents diverse skin tones, motions, and age groups, enriching our evaluation with complex real-world cases.

As illustrated in Table \ref{tab:tab1}, our model significantly outperforms the benchmark models in maintaining continuity between frames, as evidenced by the TC score, which surpasses the next best by $5.6\%$. Furthermore, achieving the highest overall metric scores across various datasets demonstrates our model's ability to capture detailed, realistic structures and ensures its robustness beyond our training dataset. Additionally, we observe an all-rounded better performance of segmented masks over bounding boxes, which is expected since detailed images of parts of a human body come in irregular shapes, and thus bounding boxes mismatch. 

Looking at the test results on the HOF dataset, we have strong evidence of DiffMVR's mightiness in capturing authentic, continuous details from a general viewpoint. 

By observing the numeric results in Table \ref{tab:tab1}, we select the Tuned-runwayml as the second-best image-level model based on its consistent performance across the metrics. We further conduct relative comparisons between DiffMVR and Tuned-runwayml, and notice a better performance of DiffMVR in all metrics, especially the structural similarity perspective. 
\begin{table*}[ht]
\centering
\scriptsize
\setlength{\tabcolsep}{2pt}
\renewcommand{\arraystretch}{0.9} 
\resizebox{\textwidth}{!}{
\begin{tabular}{l p{2cm} p{2cm} p{2cm} p{2cm} l l l l l}
\toprule
{} & \multicolumn{3}{c}{\textbf{Baby Dataset - Segmented masks}} & \multicolumn{3}{c}{\textbf{Baby Dataset - Bounding boxes}} & \multicolumn{3}{c}{\textbf{HOF Dataset - Segmented masks}} \\
{Model} & {FID $\downarrow$} & {SSIM $\uparrow$} & {TC $\downarrow$} & {FID $\downarrow$} & {SSIM $\uparrow$} & {TC $\downarrow$} & {FID $\downarrow$} & {SSIM $\uparrow$} & {TC $\downarrow$} \\
\midrule
\rowcolor{gray!30} 
DiffMVR & $\textbf{2.363}$ & $\textbf{0.898}$ & \textbf{0.395} & $2.196$ & $\textbf{0.864}$ & \textbf{0.393} & $\textbf{5.412}$ & $\textbf{0.786}$ & \textbf{0.428} \\
\cmidrule(l){1-10}
Stabilityai & $3.066 \textcolor{brightpink}{\scriptstyle \blacktriangle 29.8\%}$
 & $0.689 \textcolor{bluearrow}{\scriptstyle \blacktriangledown 23.3\%}$ & $0.428 \textcolor{brightpink}{\scriptstyle \blacktriangle 8.4\%}$ & $3.230 \textcolor{brightpink}{\scriptstyle \blacktriangle 47.0\%}$ & $0.706 \textcolor{bluearrow}{\scriptstyle \blacktriangledown 18.3\%}$ & $0.430 \textcolor{brightpink}{\scriptstyle \blacktriangle 9.4\%}$ & $6.118 \textcolor{brightpink}{\scriptstyle \blacktriangle 13.0\%}$ & $0.751 \textcolor{bluearrow}{\scriptstyle \blacktriangledown 4.5\%}$ & $0.430 \textcolor{brightpink}{\scriptstyle \blacktriangle 0.5\%}$ \\
Tuned-stabilityai & $2.782 \textcolor{brightpink}{\scriptstyle \blacktriangle 17.7\%}$ & $0.731 \textcolor{bluearrow}{\scriptstyle \blacktriangledown 18.6\%}$ & $0.417 \textcolor{brightpink}{\scriptstyle \blacktriangle 5.6\%}$ & $2.951 \textcolor{brightpink}{\scriptstyle \blacktriangle 34.4\%}$ & $0.739 \textcolor{bluearrow}{\scriptstyle \blacktriangledown 14.5\%}$ & $0.421 \textcolor{brightpink}{\scriptstyle \blacktriangle 1.1\%}$ & $6.225 \textcolor{brightpink}{\scriptstyle \blacktriangle 15.0\%}$ & $0.726 \textcolor{bluearrow}{\scriptstyle \blacktriangledown 7.6\%}$ & $0.431 \textcolor{brightpink}{\scriptstyle \blacktriangle 0.7\%}$ \\
Runwayml & $2.913 \textcolor{brightpink}{\scriptstyle \blacktriangle 33.5\%}$ & $0.749 \textcolor{bluearrow}{\scriptstyle \blacktriangledown 16.6\%}$ & $0.431 \textcolor{brightpink}{\scriptstyle \blacktriangle 9.1\%}$ & $2.931 \textcolor{brightpink}{\scriptstyle \blacktriangle 0.1\%}$ & $0.738 \textcolor{bluearrow}{\scriptstyle \blacktriangledown 14.6\%}$ & $0.433 \textcolor{brightpink}{\scriptstyle \blacktriangle 10.2\%}$ & $5.943 \textcolor{brightpink}{\scriptstyle \blacktriangle 9.8\%}$ & $0.742 \textcolor{bluearrow}{\scriptstyle \blacktriangledown 5.6\%}$ & $0.430 \textcolor{brightpink}{\scriptstyle \blacktriangle 0.5\%}$ \\
Tuned-runwayml & $2.365 \textcolor{brightpink}{\scriptstyle \blacktriangle 0.1\%}$ & $0.760 \textcolor{bluearrow}{\scriptstyle \blacktriangledown 15.4\%}$ & $0.430 \textcolor{brightpink}{\scriptstyle \blacktriangle 8.9\%}$ & $\textbf{2.126} \textcolor{bluearrow}{\scriptstyle \blacktriangledown 3.2\%}$
 & $0.745 \textcolor{bluearrow}{\scriptstyle \blacktriangledown 13.8\%}$ & $0.432 \textcolor{brightpink}{\scriptstyle \blacktriangle 9.9\%}$ & $6.109 \textcolor{brightpink}{\scriptstyle \blacktriangle 12.9\%}$ & $0.735 \textcolor{bluearrow}{\scriptstyle \blacktriangledown 6.5\%}$ & $0.434 \textcolor{brightpink}{\scriptstyle \blacktriangle 1.4\%}$ \\
LaMa & $2.940 \textcolor{brightpink}{\scriptstyle \blacktriangle 24.4\%}$ & $0.712 \textcolor{bluearrow}{\scriptstyle \blacktriangledown 20.7\%}$ & $0.456 \textcolor{brightpink}{\scriptstyle \blacktriangle 15.4\%}$ & $3.105 \textcolor{brightpink}{\scriptstyle \blacktriangle 41.4\%}$ & $0.670 \textcolor{bluearrow}{\scriptstyle \blacktriangledown 22.5\%}$ & $0.461 \textcolor{brightpink}{\scriptstyle \blacktriangle 17.3\%}$ & $7.025 \textcolor{brightpink}{\scriptstyle \blacktriangle 29.8\%}$ & $0.731 \textcolor{bluearrow}{\scriptstyle \blacktriangledown 7.0\%}$ & $0.456 \textcolor{brightpink}{\scriptstyle \blacktriangle 6.5\%}$ \\
FGVI & $2.850 \textcolor{brightpink}{\scriptstyle \blacktriangle 20.6\%}$ & $0.832 \textcolor{bluearrow}{\scriptstyle \blacktriangledown 7.3\%}$ & $0.420 \textcolor{brightpink}{\scriptstyle \blacktriangle 6.3\%}$ & $2.902 \textcolor{brightpink}{\scriptstyle \blacktriangle 32.1\%}$ & $0.831 \textcolor{bluearrow}{\scriptstyle \blacktriangledown 3.8\%}$ & $0.423 \textcolor{brightpink}{\scriptstyle \blacktriangle 7.6\%}$ & $6.299 \textcolor{brightpink}{\scriptstyle \blacktriangle 16.4\%}$ & $0.747 \textcolor{bluearrow}{\scriptstyle \blacktriangledown 5.0\%}$ & $0.431 \textcolor{brightpink}{\scriptstyle \blacktriangle 0.7\%}$ \\
PVI & $2.773 \textcolor{brightpink}{\scriptstyle \blacktriangle 17.4\%}$ & $0.844 \textcolor{bluearrow}{\scriptstyle \blacktriangledown 6.0\%}$ & $\textbf{0.395}$ & $2.858 \textcolor{brightpink}{\scriptstyle \blacktriangle 30.1\%}$ & $0.836 \textcolor{bluearrow}{\scriptstyle \blacktriangledown 3.2\%}$ & $0.396 \textcolor{brightpink}{\scriptstyle \blacktriangle 0.8\%}$ & $6.345 \textcolor{brightpink}{\scriptstyle \blacktriangle 17.2\%}$ & $0.762 \textcolor{bluearrow}{\scriptstyle \blacktriangledown 3.1\%}$ & $0.429 \textcolor{brightpink}{\scriptstyle \blacktriangle 0.2\%}$ \\
\cmidrule(l){1-10}
\rowcolor{gray!10} 
Gap & $+0.08\%$ & $+18.16\%$ & $+8.14\%$ & $-3.29\%$ & $+15.97\%$ & $+9.03\%$ & $+11.41\%$ & $+6.94\%$ & $+1.38\%$ \\
\rowcolor{gray!10} 
Gap between Masks & $+7.60\%$ & $+3.94\%$ & $-0.51\%$ & $---$ & $---$ & $---$ & $---$ & $---$ & $---$ \\
\bottomrule
\end{tabular}
}
\caption{Quantitative results comparing different models using FID, SSIM, and TC metrics on \textbf{frame-level} for the Baby and HOF datasets. The HOF dataset is used for proving the generality of DiffMVR. Dash means the value is undefined. The $\textcolor{brightpink}{\blacktriangle}$/$\textcolor{bluearrow}{\blacktriangledown}$ indicates a relative increase/decrease in metric score compared to DiffMVR. \textit{Gap} refers to the extent by which DiffMVR outperforms (+) or is outperformed by (-) the second-best model (\textit{Tuned-runwayml}). \textit{Gap between Masks} refers to the extent by which segmented masks outperforms (+) or is outperformed by (-) the bounding boxes, both within DiffMVR model.}
\label{tab:tab1}
\end{table*}

\subsubsection{Video-level}
DiffMVR achieves the best scores for both segmented masks and bounding boxes, as shown in Table \ref{tab:tabv}. Apparently, image-based models suffer heavily from inconsistent object attributes and discontinuity. Based on the test scores, we select PVI as the second-best model. PVI has a superior ability in constructing spatial-similar videos. However, when it comes to the comparison of the FVD score, which is a combined metric that evaluates the integrated performance on structural similarity and temporal coherence, DiffMVR stands out. DiffMVR persistently achieves the best SSIM and TC scores, this substantial margin highlights DiffMVR's effectiveness in managing both the larger Baby dataset and the much smaller HOF dataset. To further demonstrate the overall robustness of DiffMVR, we show the visualized results in the following section.

\begin{table*}[ht]
\centering
\scriptsize
\setlength{\tabcolsep}{6pt} 
\renewcommand{\arraystretch}{0.9} 
\begin{tabular}{lllllllll}
\toprule
{} & \multicolumn{4}{c}{\textbf{Baby Dataset - Segmented masks}} & \multicolumn{4}{c}{\textbf{Baby Dataset - Bounding boxes}} \\
{Model} & {\(\bar{\text{FID}}\) $\downarrow$} & {\(\bar{\text{SSIM}}\) $\uparrow$} & {\(\bar{\text{TC}}\) $\downarrow$} & {\({\text{FVD}}\) $\downarrow$} & {\(\bar{\text{FID}}\) $\downarrow$} & {\(\bar{\text{SSIM}}\) $\uparrow$} & {\(\bar{\text{TC}}\) $\downarrow$} & {\({\text{FVD}}\) $\downarrow$} \\
\midrule
\rowcolor{gray!30}
DiffMVR & $2.095$ & $\textbf{0.908}$ & \textbf{0.338} &  $\textbf{48.05}$ & $2.119$ & $\textbf{0.880}$ & \textbf{0.341} & $\textbf{50.47}$ \\
\cmidrule(l){1-9}
Stabilityai & $2.406 \textcolor{brightpink}{\scriptstyle \blacktriangle 14.8\%}$ & $0.738 \textcolor{bluearrow}{\scriptstyle \blacktriangledown 18.7\%}$ & $0.421 \textcolor{brightpink}{\scriptstyle \blacktriangle 24.6\%}$ & $73.94 \textcolor{brightpink}{\scriptstyle \blacktriangle 53.9\%}$ & $2.497 \textcolor{brightpink}{\scriptstyle \blacktriangle 17.8\%}$ & $0.736 \textcolor{bluearrow}{\scriptstyle \blacktriangledown 16.4\%}$ & $0.427 \textcolor{brightpink}{\scriptstyle \blacktriangle 25.2\%}$ & $74.39 \textcolor{brightpink}{\scriptstyle \blacktriangle 47.4\%}$ \\
Tuned-stabilityai & $2.352 \textcolor{brightpink}{\scriptstyle \blacktriangle 12.3\%}$ & $0.756 \textcolor{bluearrow}{\scriptstyle \blacktriangledown 16.7\%}$ & $0.398 \textcolor{brightpink}{\scriptstyle \blacktriangle 17.8\%}$ & $71.28 \textcolor{brightpink}{\scriptstyle \blacktriangle 48.3\%}$ & $2.414 \textcolor{brightpink}{\scriptstyle \blacktriangle 13.9\%}$ & $0.747 \textcolor{bluearrow}{\scriptstyle \blacktriangledown 15.1\%}$ & $0.401 \textcolor{brightpink}{\scriptstyle \blacktriangle 17.6\%}$ & $73.06 \textcolor{brightpink}{\scriptstyle \blacktriangle 44.8\%}$  \\
Runwayml & $2.410 \textcolor{brightpink}{\scriptstyle \blacktriangle 15.0\%}$ & $0.759 \textcolor{bluearrow}{\scriptstyle \blacktriangledown 16.4\%}$ & $0.423 \textcolor{brightpink}{\scriptstyle \blacktriangle 25.1\%}$ & $73.02 \textcolor{brightpink}{\scriptstyle \blacktriangle 52.0\%}$ & $2.463 \textcolor{brightpink}{\scriptstyle \blacktriangle 16.2\%}$ & $0.748 \textcolor{bluearrow}{\scriptstyle \blacktriangledown 15.0\%}$ & $0.408 \textcolor{brightpink}{\scriptstyle \blacktriangle 19.6\%}$ & $73.85 \textcolor{brightpink}{\scriptstyle \blacktriangle 46.3\%}$ \\
Tuned-runwayml & $2.247 \textcolor{brightpink}{\scriptstyle \blacktriangle 7.3\%}$ & $0.763 \textcolor{bluearrow}{\scriptstyle \blacktriangledown 16.0\%}$ & $0.417 \textcolor{brightpink}{\scriptstyle \blacktriangle 23.4\%}$ & $70.86 \textcolor{brightpink}{\scriptstyle \blacktriangle 47.5\%}$ & $2.229 \textcolor{brightpink}{\scriptstyle \blacktriangle 5.2\%}$ & $0.749 \textcolor{bluearrow}{\scriptstyle \blacktriangledown 14.9\%}$ & $0.420 \textcolor{brightpink}{\scriptstyle \blacktriangle 23.2\%}$ & $72.27 \textcolor{brightpink}{\scriptstyle \blacktriangle 43.2\%}$\\
LaMa & $2.933 \textcolor{brightpink}{\scriptstyle \blacktriangle 40.0\%}$ & $0.720 \textcolor{bluearrow}{\scriptstyle \blacktriangledown 20.7\%}$ & $0.454 \textcolor{brightpink}{\scriptstyle \blacktriangle 34.3\%}$ & $77.95 \textcolor{brightpink}{\scriptstyle \blacktriangle 62.2\%}$ & $3.195 \textcolor{brightpink}{\scriptstyle \blacktriangle 50.8\%}$ & $0.695 \textcolor{bluearrow}{\scriptstyle \blacktriangledown 21.0\%}$ & $0.455 \textcolor{brightpink}{\scriptstyle \blacktriangle 33.4\%}$ & $78.12 \textcolor{brightpink}{\scriptstyle \blacktriangle 54.8\%}$  \\
FGVI & $2.115 \textcolor{brightpink}{\scriptstyle \blacktriangle 1.0\%}$ & $0.849 \textcolor{bluearrow}{\scriptstyle \blacktriangledown 6.5\%}$ & $0.350 \textcolor{brightpink}{\scriptstyle \blacktriangle 3.6\%}$ & $52.75 \textcolor{brightpink}{\scriptstyle \blacktriangle 9.8\%}$ & $2.142 \textcolor{brightpink}{\scriptstyle \blacktriangle 1.1\%}$ & $0.845 \textcolor{bluearrow}{\scriptstyle \blacktriangledown 4.0\%}$ & $0.351 \textcolor{brightpink}{\scriptstyle \blacktriangle 2.9\%}$ & $55.60 \textcolor{brightpink}{\scriptstyle \blacktriangle 10.2\%}$  \\
PVI & $\textbf{2.062} \textcolor{bluearrow}{\scriptstyle \blacktriangledown 1.6\%}$ & $0.894 \textcolor{bluearrow}{\scriptstyle \blacktriangledown 1.5\%}$ & $0.339 \textcolor{brightpink}{\scriptstyle \blacktriangle 0.3\%}$ & $48.92 \textcolor{brightpink}{\scriptstyle \blacktriangle 1.8\%}$ & $\textbf{2.105} \textcolor{bluearrow}{\scriptstyle \blacktriangledown 0.7\%}$ & $0.860 \textcolor{bluearrow}{\scriptstyle \blacktriangledown 2.3\%}$ & $0.346 \textcolor{brightpink}{\scriptstyle \blacktriangle 1.5\%}$ & $51.04 \textcolor{brightpink}{\scriptstyle \blacktriangle 1.1\%}$ \\
\cmidrule(l){1-9} 
\rowcolor{gray!10}
Gap & $-1.60\%$ & $+1.25\%$ & $+0.29\%$ & $+1.78\%$ & $-0.67\%$ & $+2.33\%$ & $+1.45\%$ & $+1.12\%$  \\
\rowcolor{gray!10}
Gap between Masks & $+1.13\%$ & $+3.18\%$ & $+0.88\%$ & $+4.79\%$ & $---$ & $---$ & $---$ & $---$  \\
\bottomrule
\end{tabular}
\caption{Quantitative results comparing different models using FID, SSIM, TC, and FVD metrics on \textbf{video-level} for the Baby dataset. We have \textit{PVI} as the second-best model, for it has the majority of the second placement in metric values. See the caption of Table\ref{tab:tab1} for other explanations.}
\label{tab:tabv}
\end{table*}

\subsection{Qualitative Results}
\begin{figure}[ht]
\centering
\hspace*{-15pt}
\setlength{\tabcolsep}{0.9pt} 
\renewcommand{\arraystretch}{4.0} 
\begin{tabular}{c c c c c c}
    \scriptsize \textbf{Baby 1} & \scriptsize \textbf{Baby 2} & \scriptsize \textbf{Baby 3} & \scriptsize \textbf{Baby 4} & \scriptsize \textbf{Baby 5} & \scriptsize \textbf{Baby 6} \\
    \includegraphics[width=0.075\textwidth]{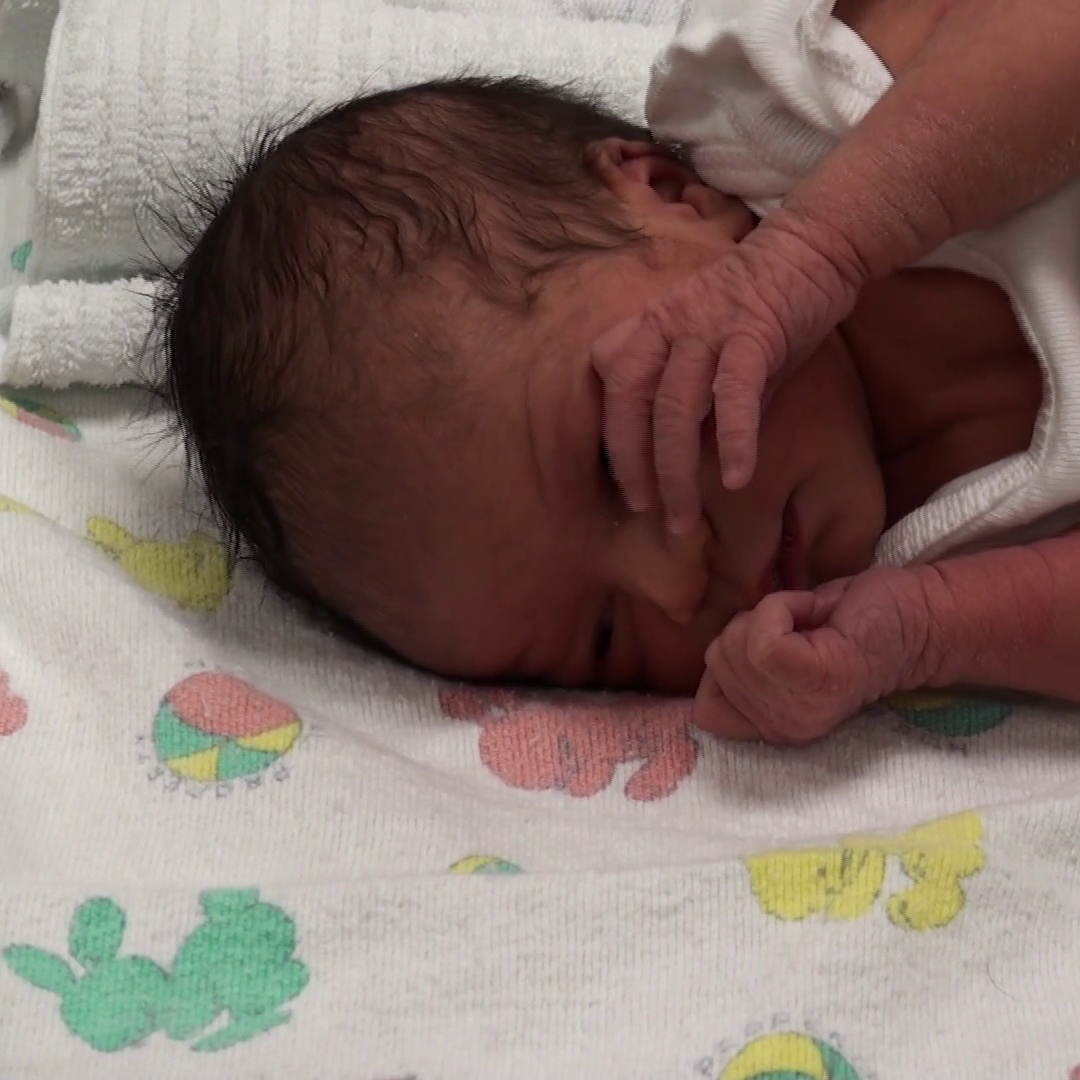} &
    \includegraphics[width=0.075\textwidth]{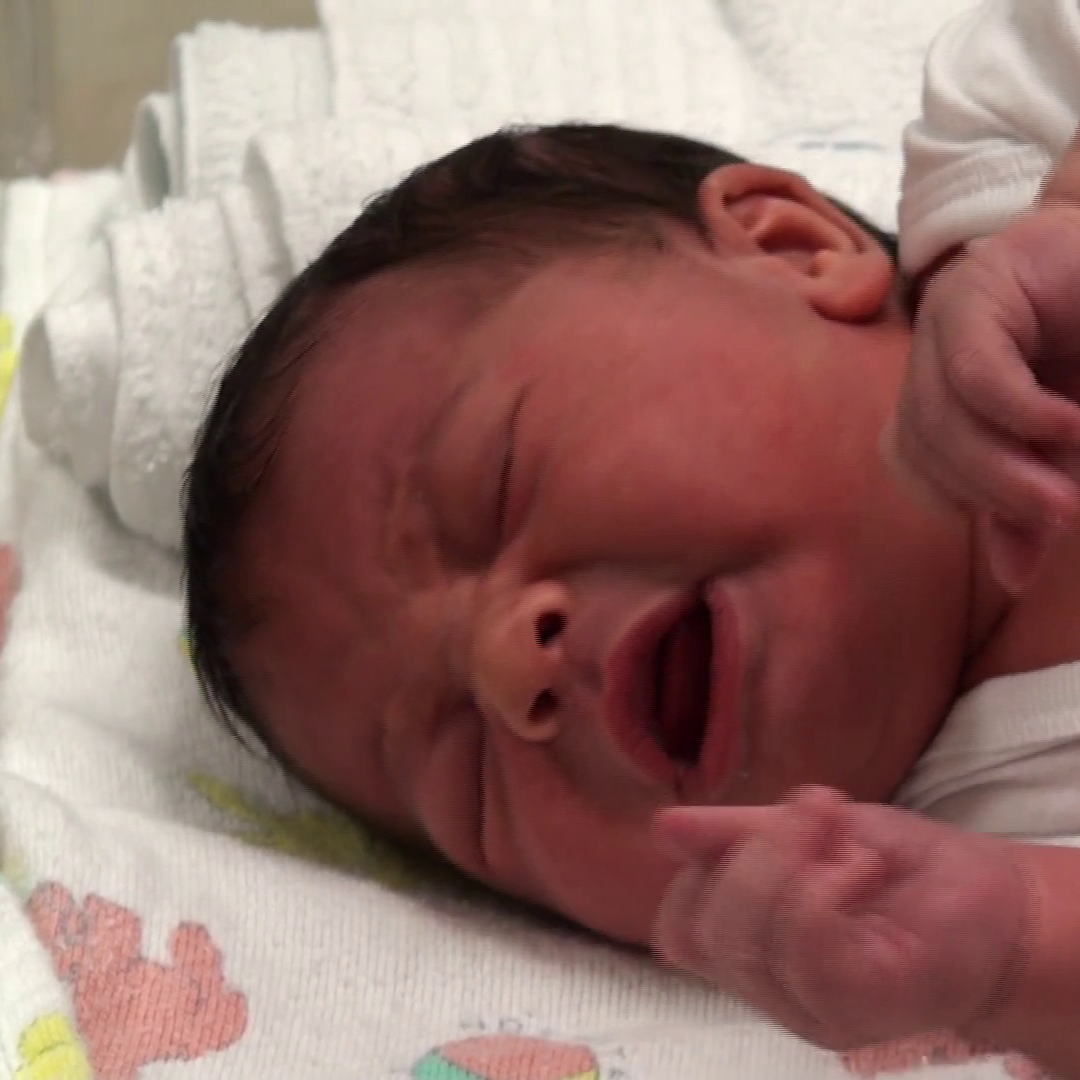} &
    \includegraphics[width=0.075\textwidth]{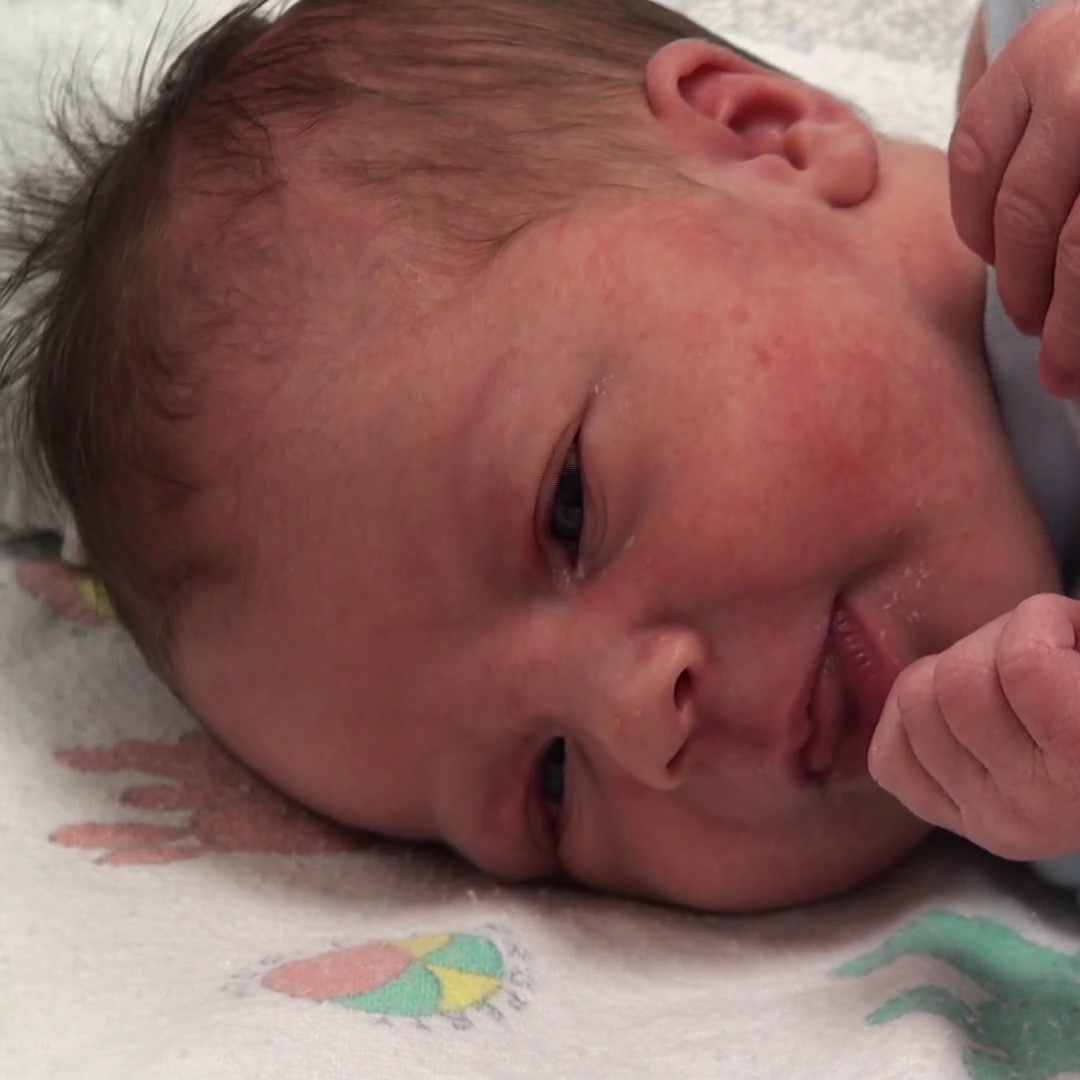} &
    \includegraphics[width=0.075\textwidth]{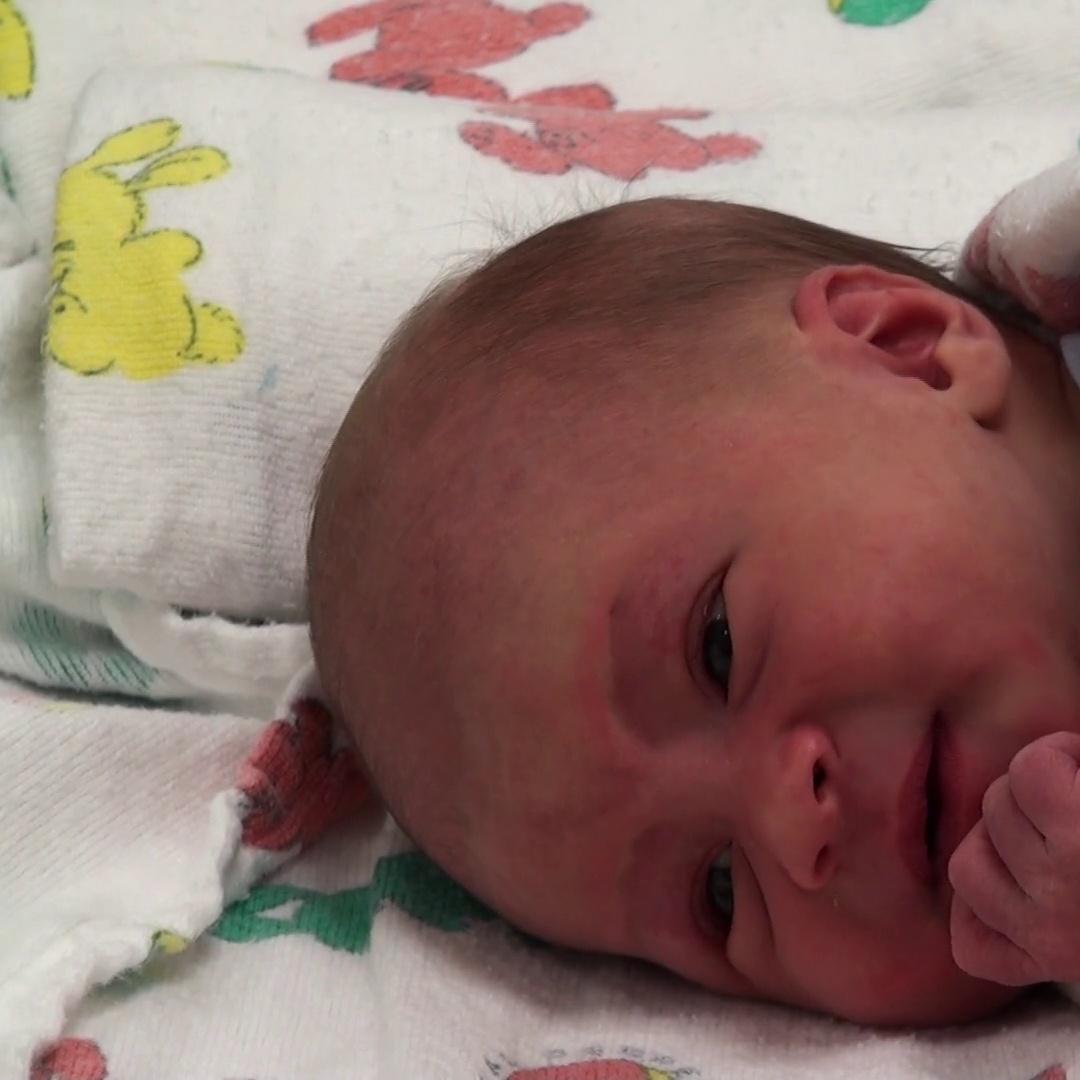} &
    \includegraphics[width=0.075\textwidth]{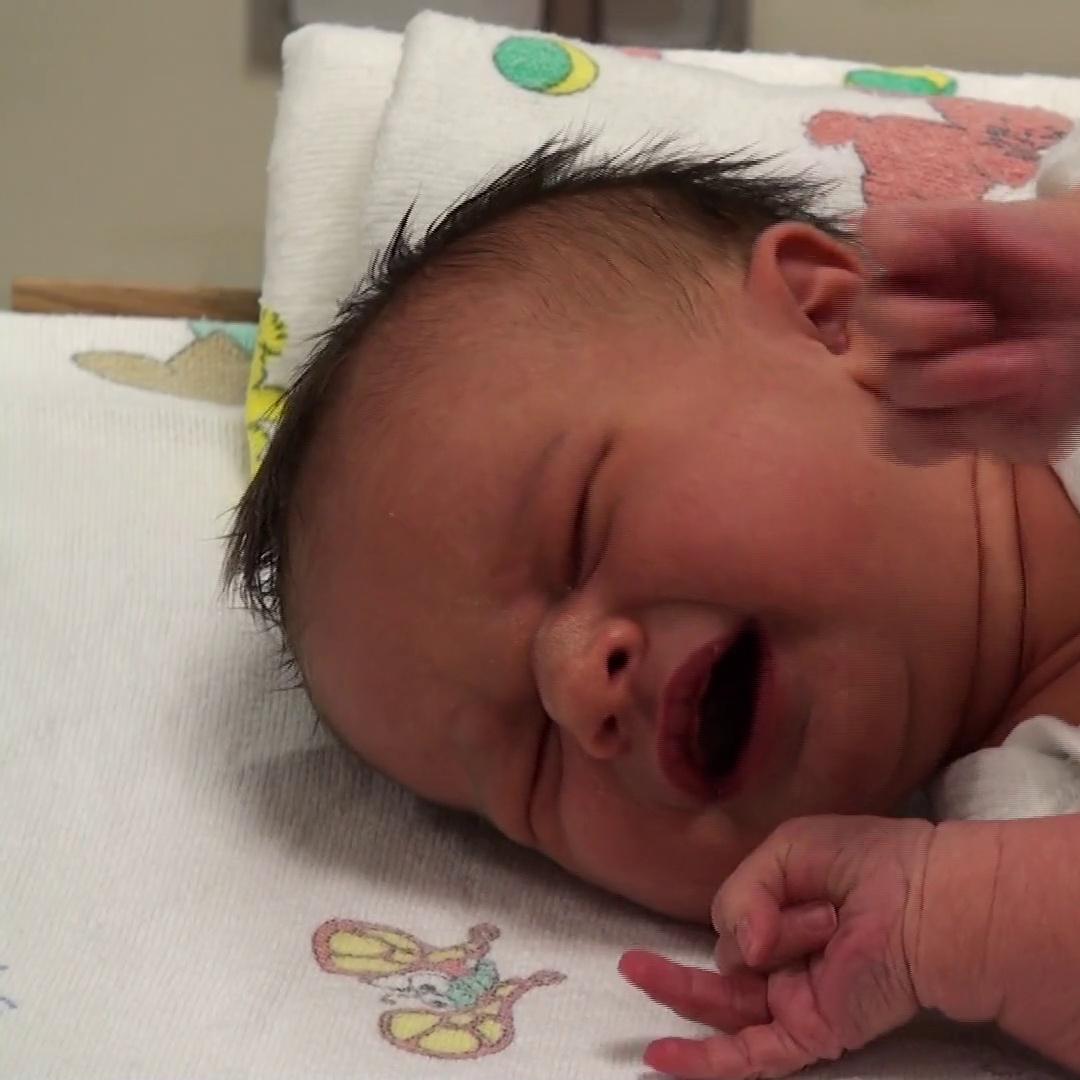} &
    \includegraphics[width=0.075\textwidth]{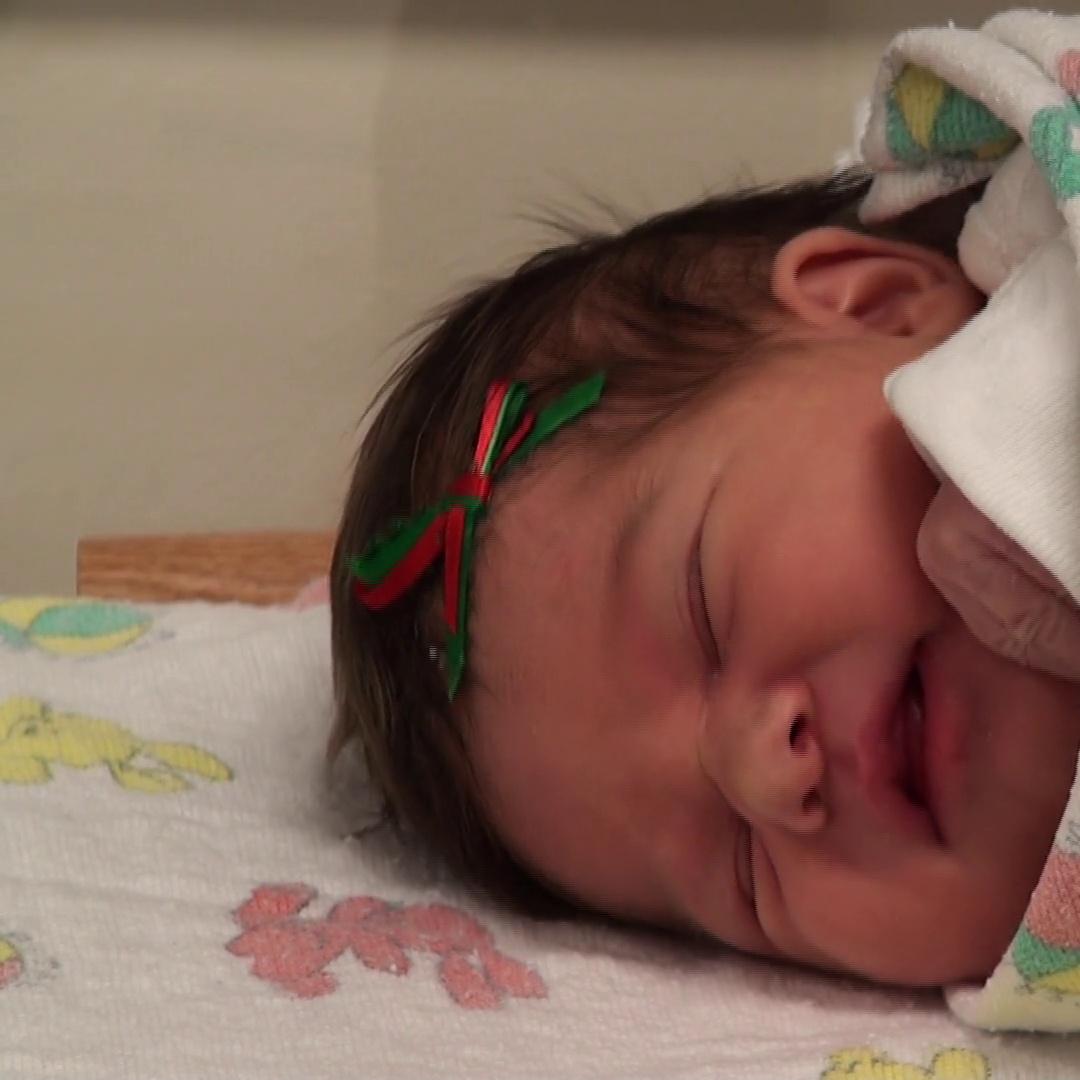} \\[-0.7em]

    \includegraphics[width=0.075\textwidth]{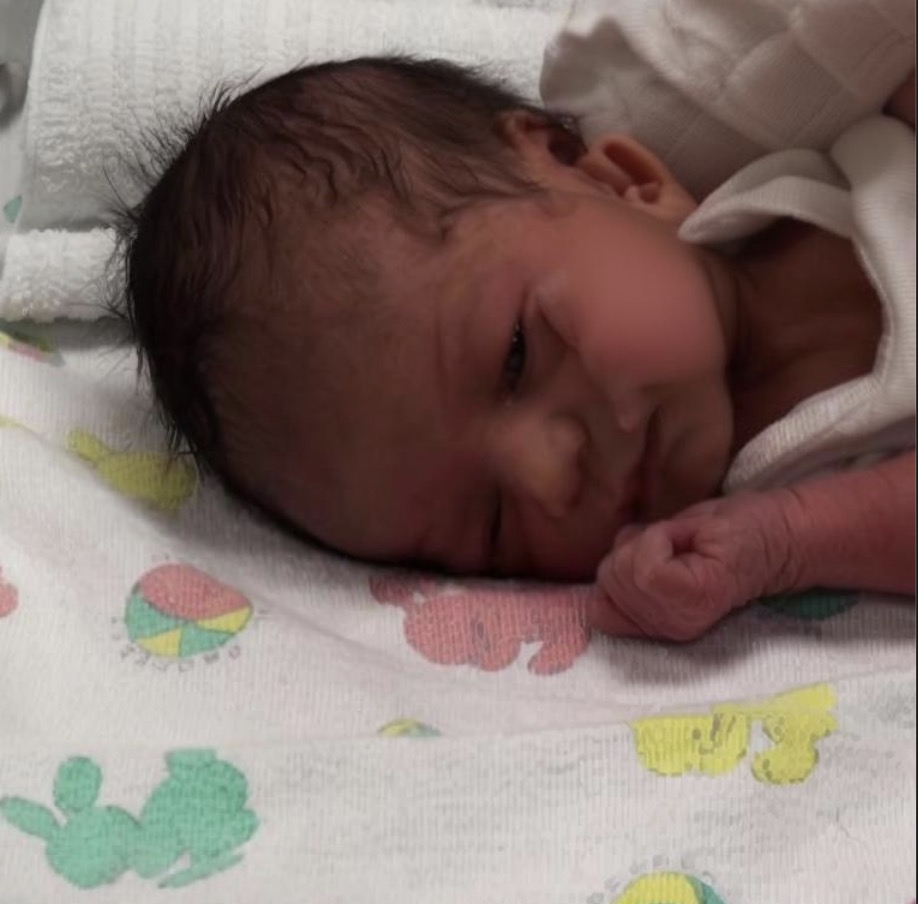} &
    \includegraphics[width=0.075\textwidth]{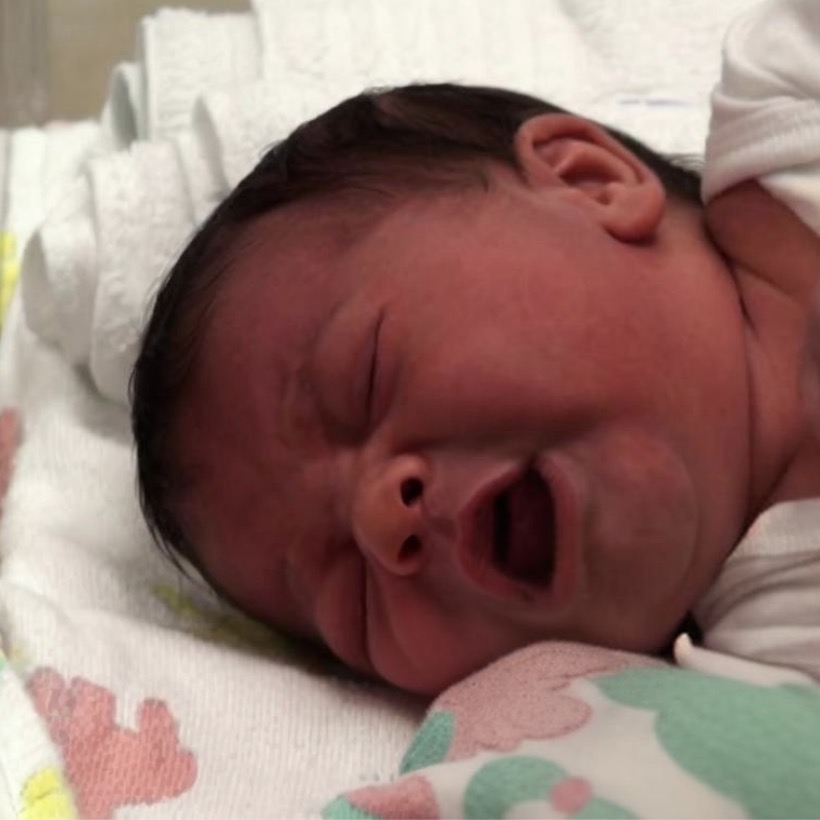} &
    \includegraphics[width=0.075\textwidth]{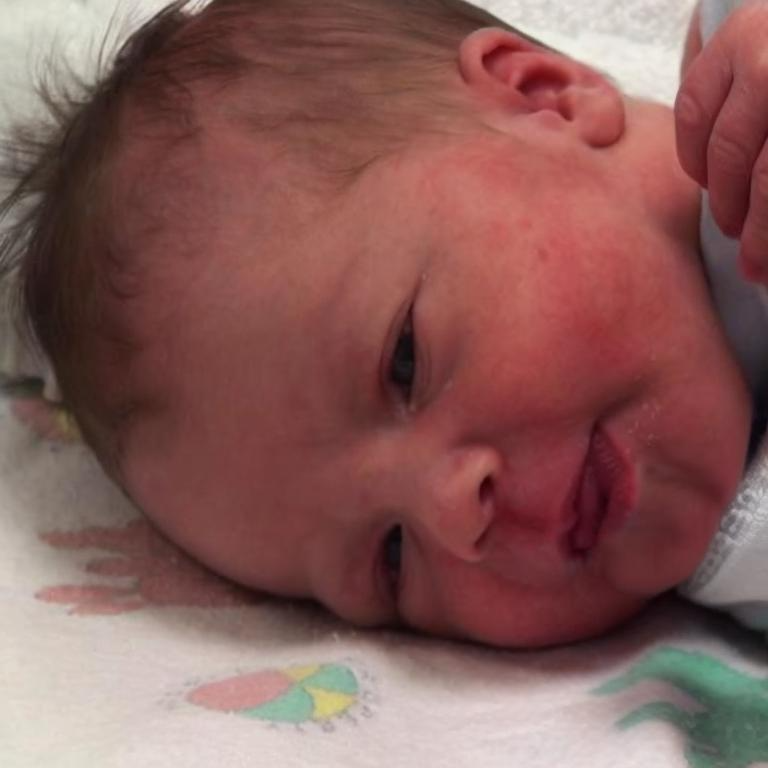} &
    \includegraphics[width=0.075\textwidth]{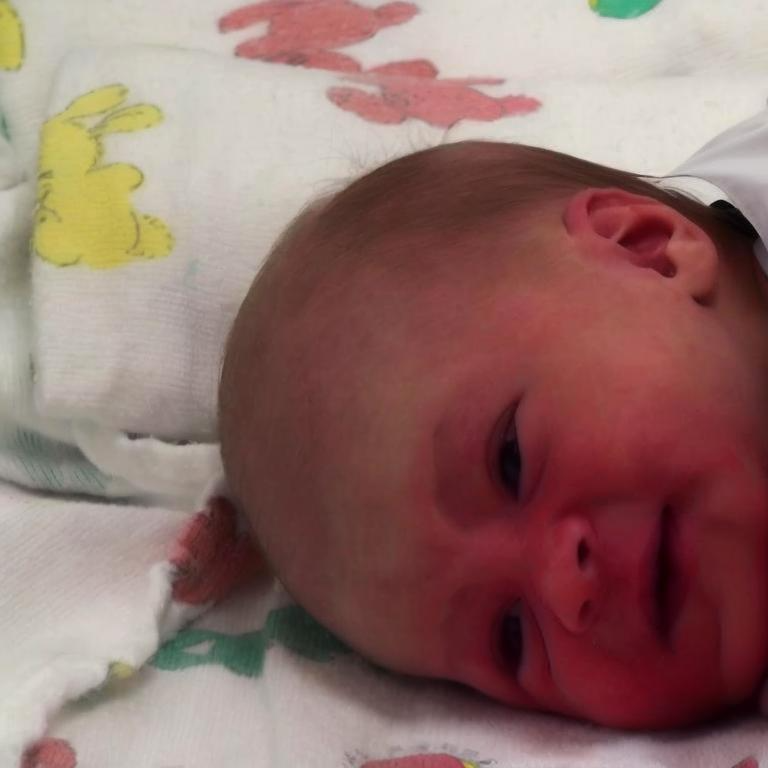} &
    \includegraphics[width=0.075\textwidth]{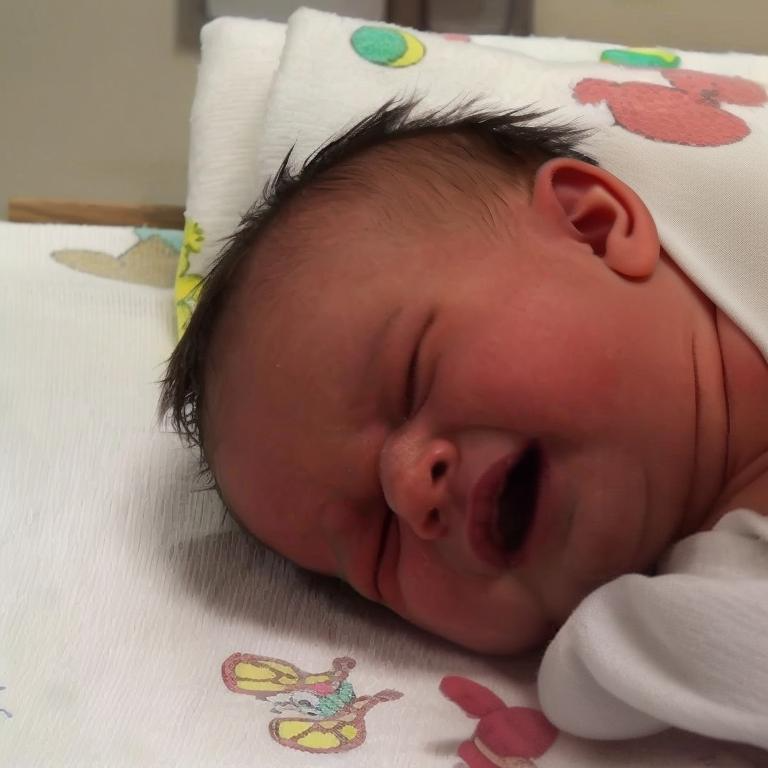} &
    \includegraphics[width=0.075\textwidth]{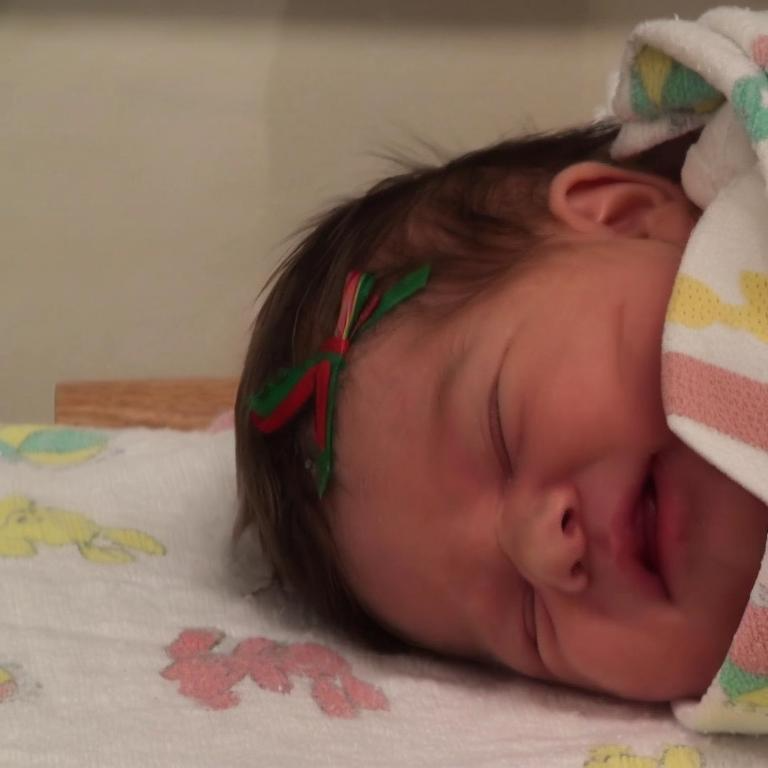} \\[-0.7em]

    \includegraphics[width=0.075\textwidth]{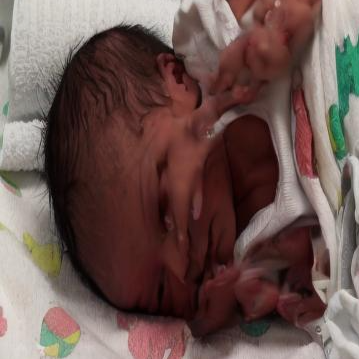} &
    \includegraphics[width=0.075\textwidth]{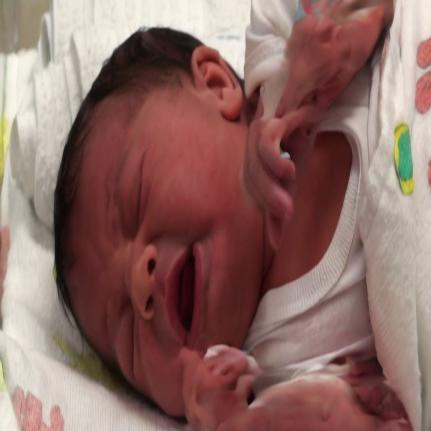} &
    \includegraphics[width=0.075\textwidth]{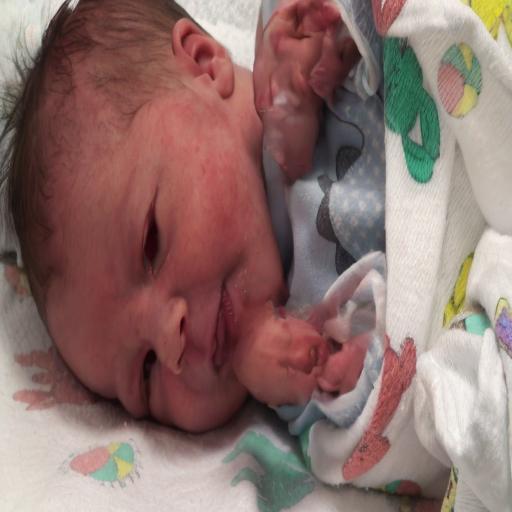} &
    \includegraphics[width=0.075\textwidth]{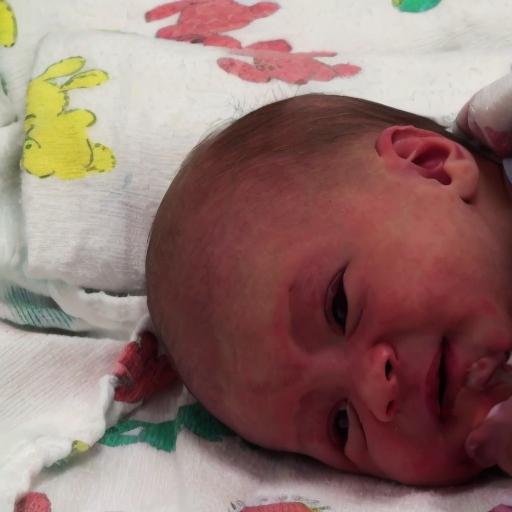} &
    \includegraphics[width=0.075\textwidth]{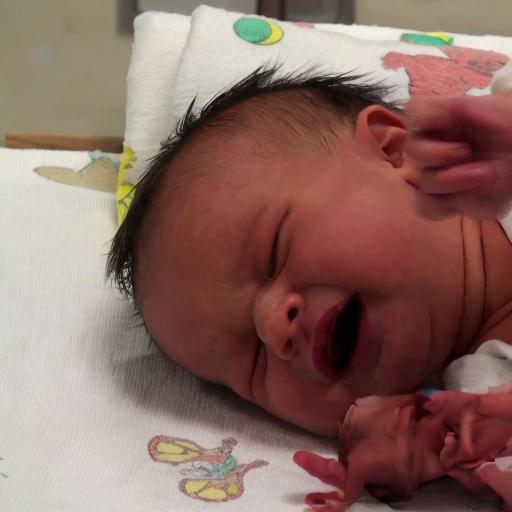} &
    \includegraphics[width=0.075\textwidth]{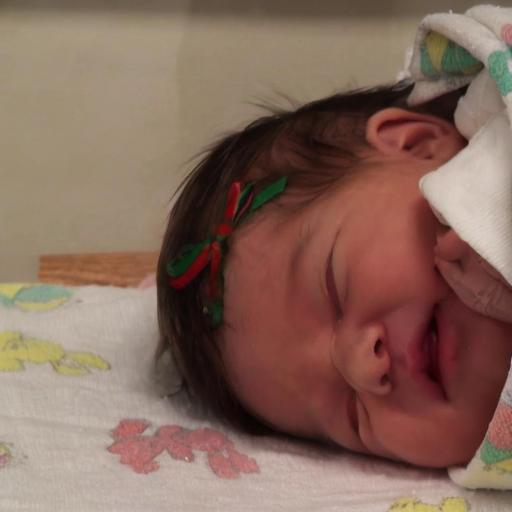} \\[-0.7em]

    \includegraphics[width=0.075\textwidth]{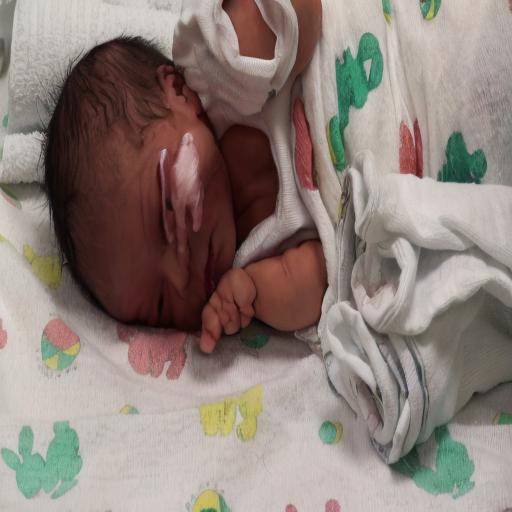} &
    \includegraphics[width=0.075\textwidth]{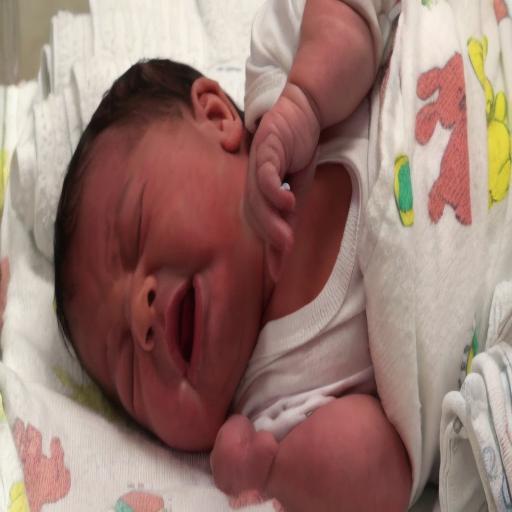} &
    \includegraphics[width=0.075\textwidth]{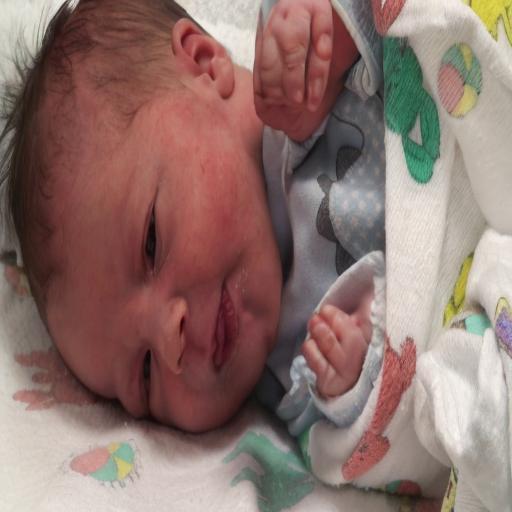} &
    \includegraphics[width=0.075\textwidth]{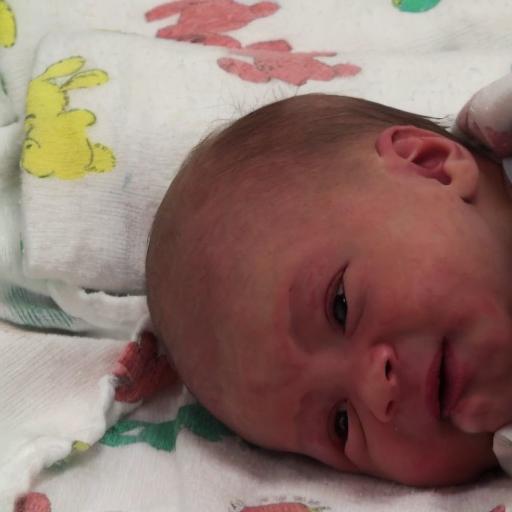} &
    \includegraphics[width=0.075\textwidth]{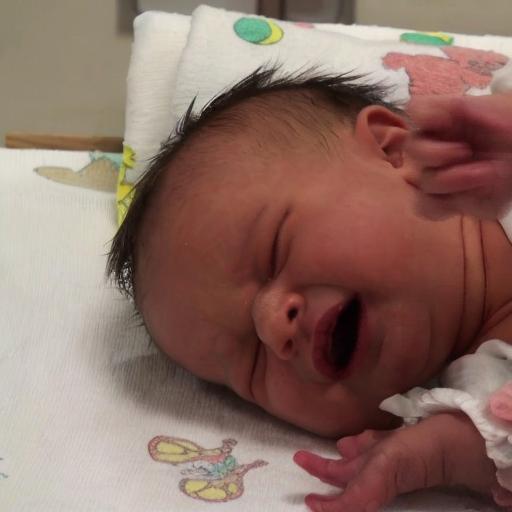} &
    \includegraphics[width=0.075\textwidth]{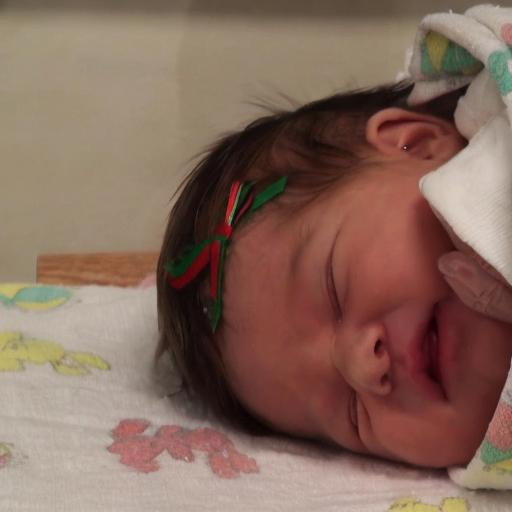} \\
\end{tabular}

\caption{Qualitative comparison of DiffMVR with the benchmarked models on the Baby dataset, including \textit{pain}, \textit{move}, and \textit{rest} babies. Row $1$ displays inputs from the video sources at the $5^{th}$ second, leveraging segmented masks. The content is copyrighted and reprinted with permission. Rows $2,3,4$ show inpainting results applying DiffMVR trained on segmented masks, with guide $1$ from the $4^{th}$ second of videos; Tuned-runwayml, using text prompt ``\textit{remove hands};'' Tuned-stabilityai, using text prompt ``\textit{remove hands}, '' respectively.}
\label{fig:qual1}
\end{figure}

To demonstrate the efficacy of our approach, we provide qualitative comparisons across videos with varying durations and complexities of masking. Figure \ref{fig:qual1} offers a side-by-side comparison of original and inpainted video frames, illustrating the capabilities of our method against baseline techniques. 

\begin{figure}[ht]
\centering
\setlength{\tabcolsep}{1pt} 
\renewcommand{\arraystretch}{0.8} 
\begin{tabular}{p{1cm} p{1cm} p{1cm} p{1cm} p{1cm} | p{1cm} p{1cm}}

    \includegraphics[width=0.05\textwidth]{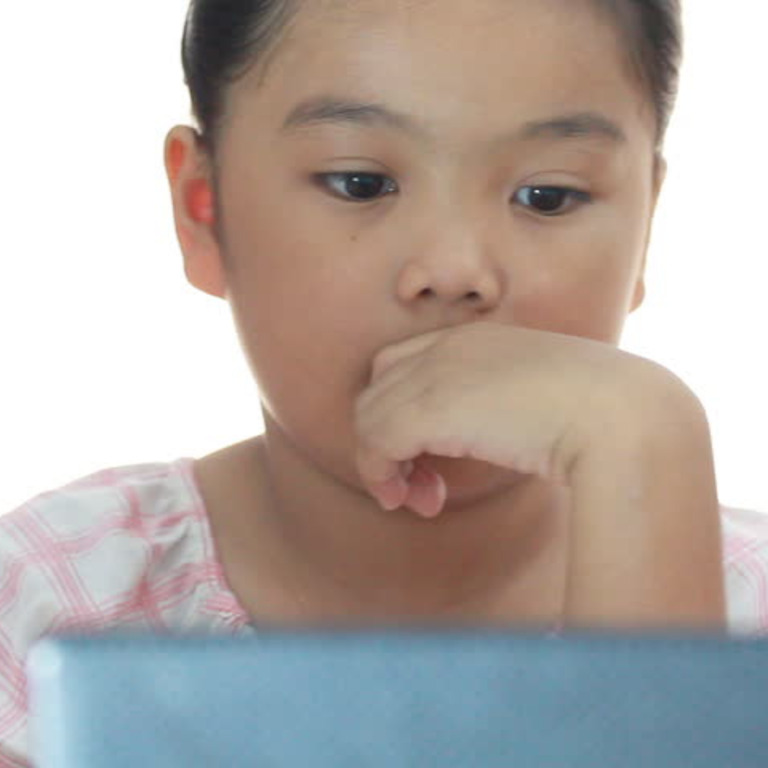} &
    \includegraphics[width=0.05\textwidth]{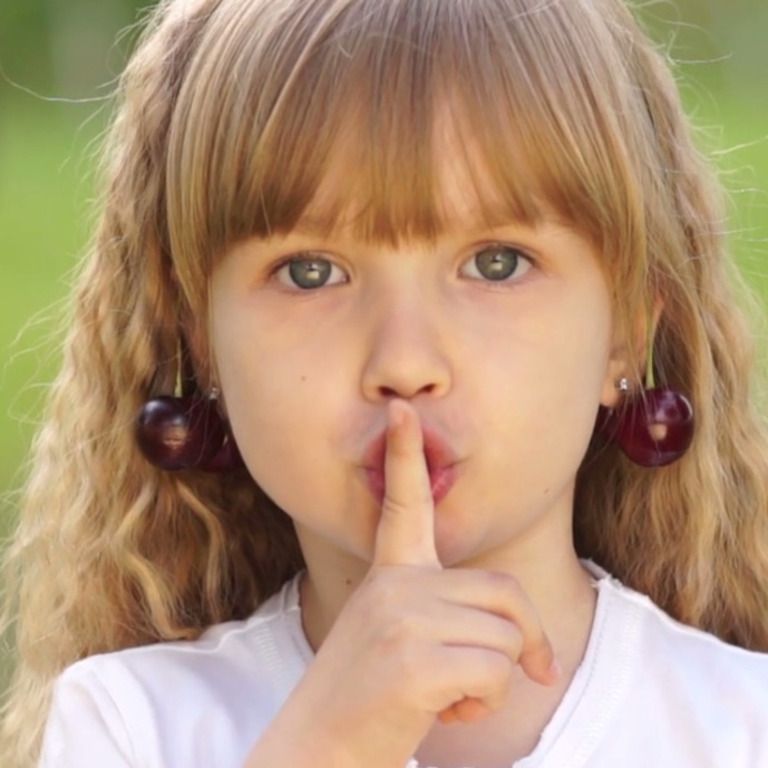} &
    \includegraphics[width=0.05\textwidth]{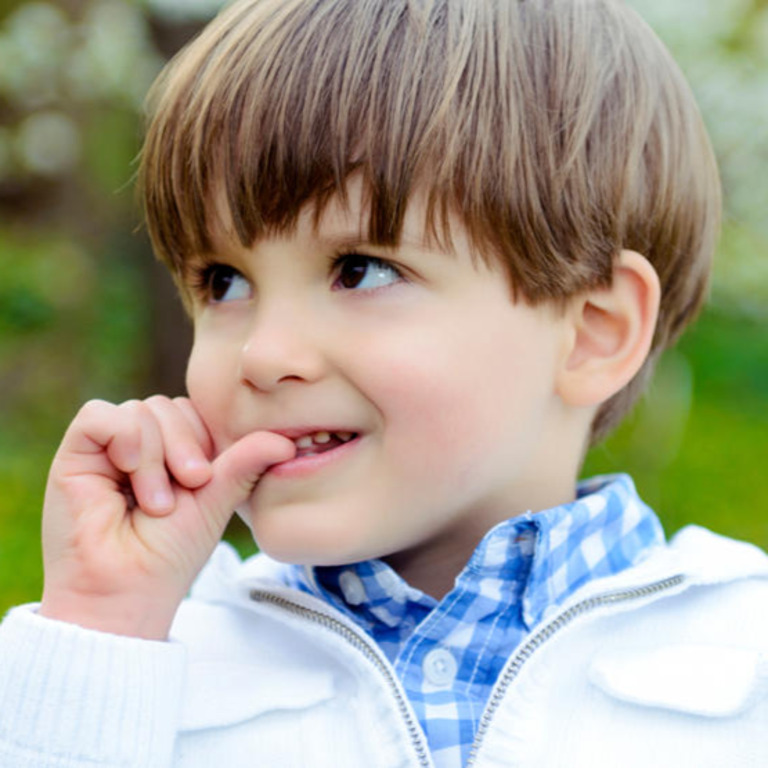} &
    \includegraphics[width=0.05\textwidth]{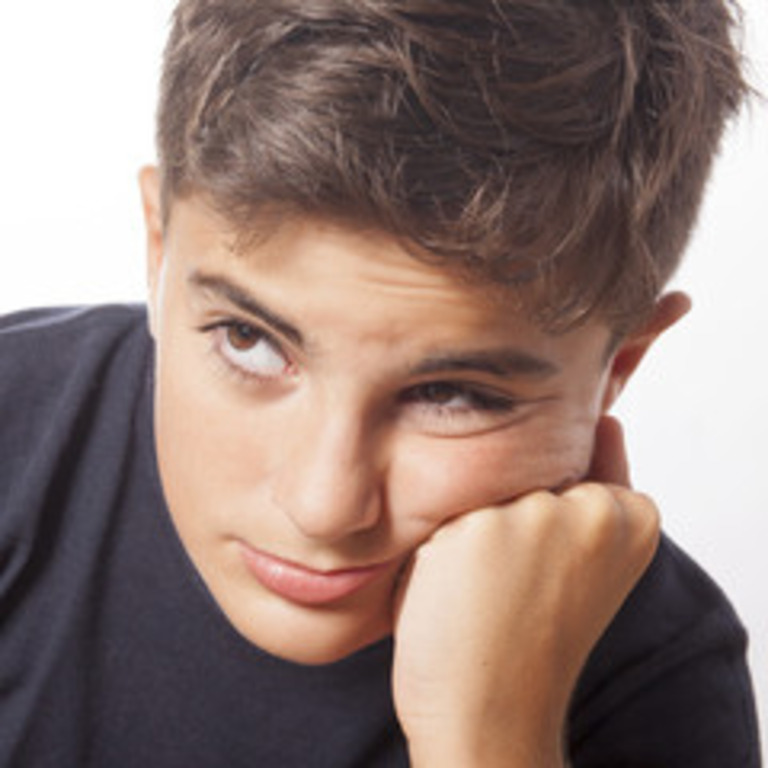} &
    \includegraphics[width=0.05\textwidth]{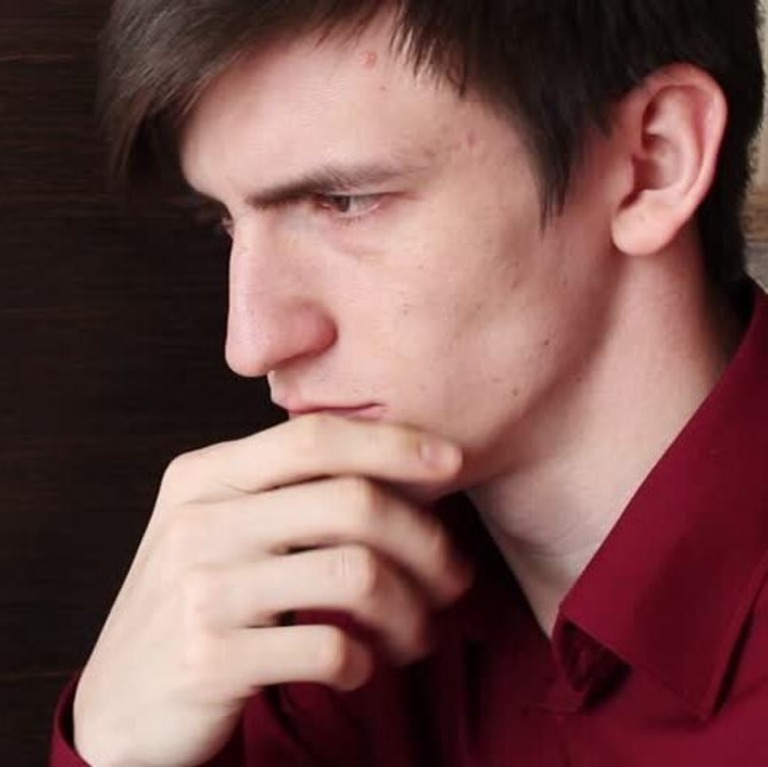} &
    \includegraphics[width=0.05\textwidth]{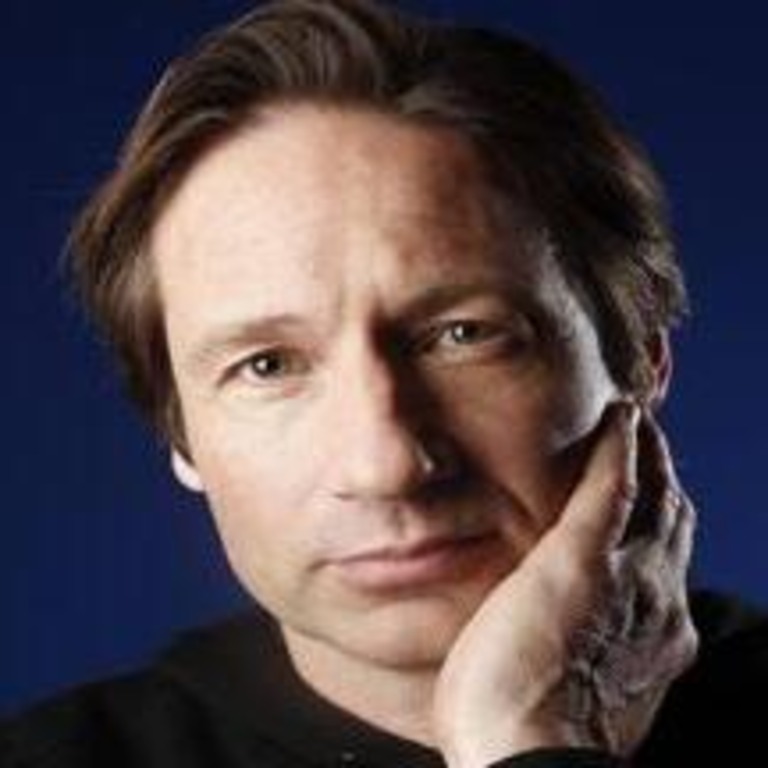} &
    \includegraphics[width=0.05\textwidth]{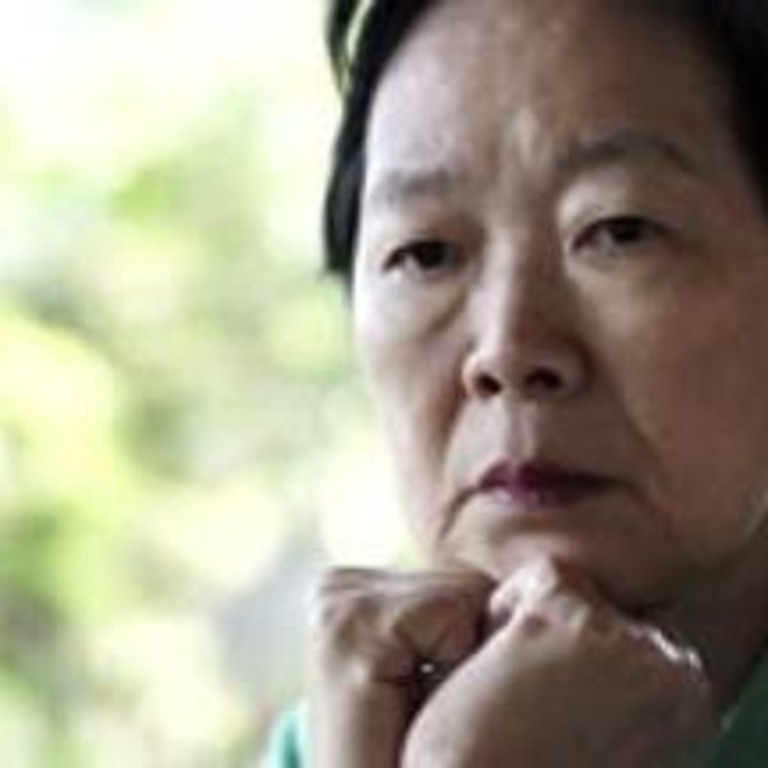} \\

    \includegraphics[width=0.05\textwidth]{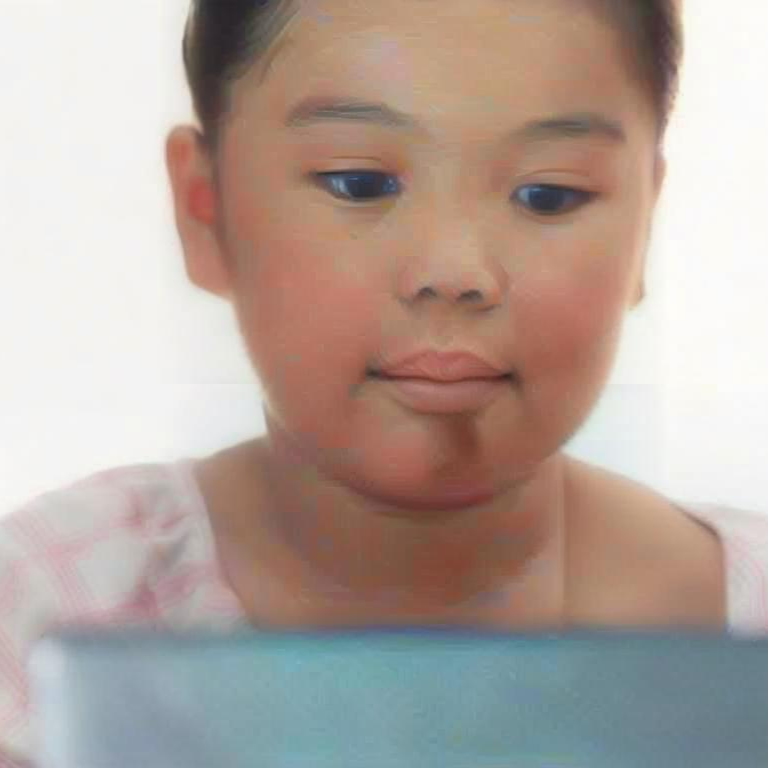} &
    \includegraphics[width=0.05\textwidth]{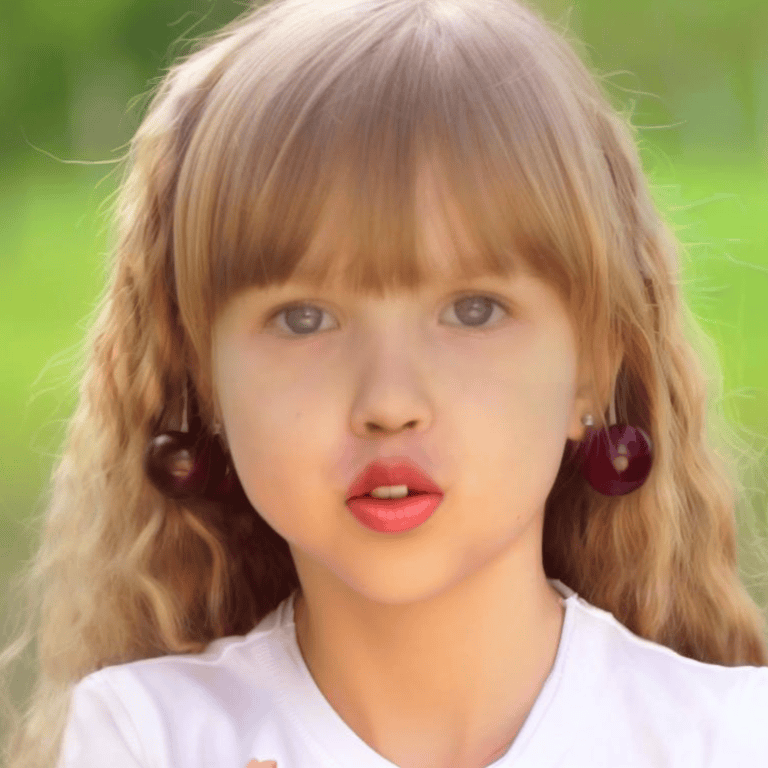} &
    \includegraphics[width=0.05\textwidth]{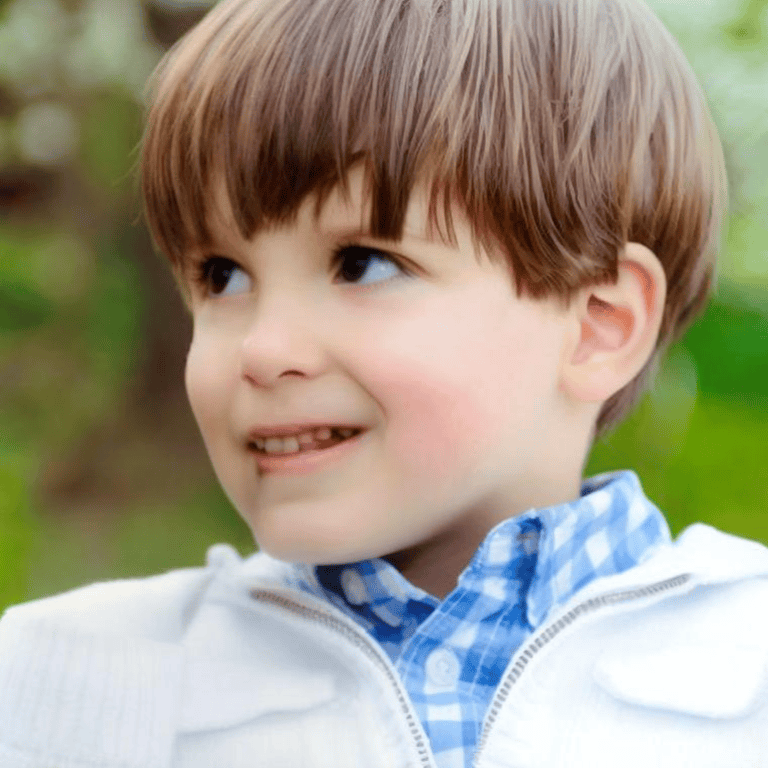} &
    \includegraphics[width=0.05\textwidth]{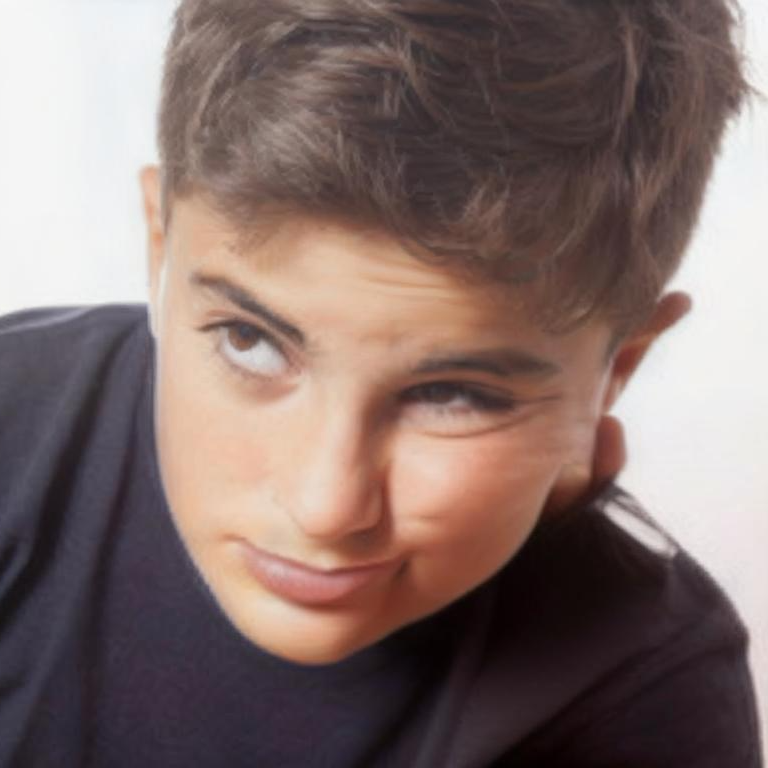} &
    \includegraphics[width=0.05\textwidth]{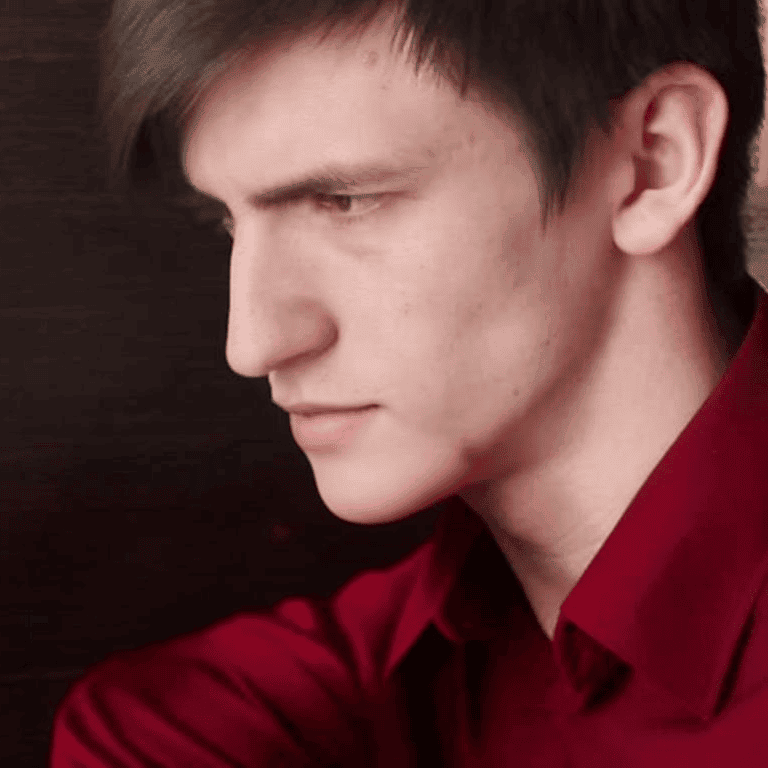} &
    \includegraphics[width=0.05\textwidth]{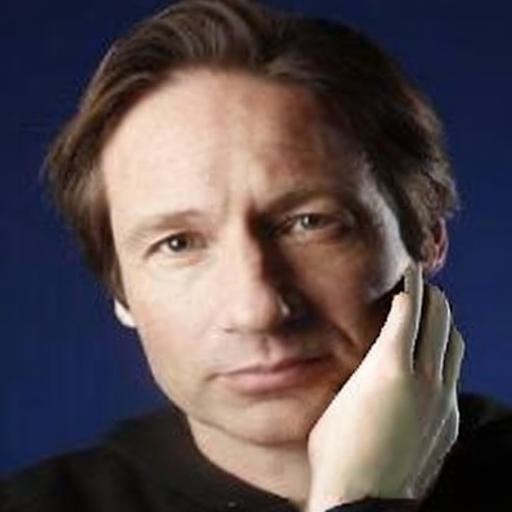} &
    \includegraphics[width=0.05\textwidth]{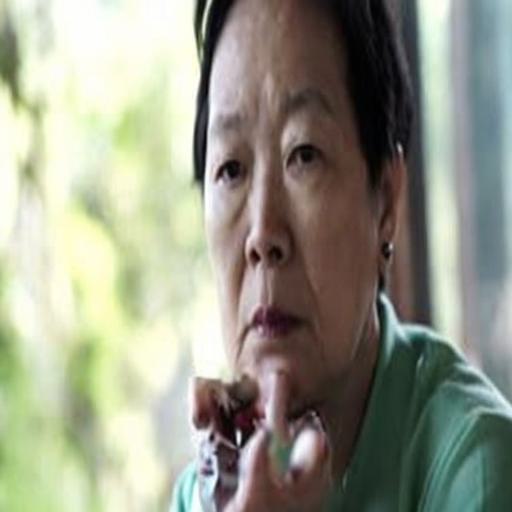} \\
\end{tabular}

\caption{Occlusion removal and face restore results on the HOF Dataset~\cite{inproceedings} applying DiffMVR. The left shows good inpaint results, and the right has some bad results. Bad could mean occlusion removal failure, restored contents incompatible with the original object, and the mask area not seamlessly connecting with the unchanged regions.}
\label{fig:half2}
\end{figure}

Further displaying the robust performance of DiffMVR, Figure \ref{fig:half2} highlights the model's ability to accurately restore dynamic scenes on out-of-distribution images. It can efficaciously reconstruct movements within the scene. Furthermore, in-distribution results in Figure \ref{fig:qual1} prove that DiffMVR is the only model that meets all the following demands: it achieves a smooth fusion between inpainted and unmasked regions, it removes obstructions, and it accurately restores the baby’s specific facial features - rather than incorrectly substituting with random body parts. Moreover, it preserves background integrity and maintains content consistency throughout. In contrast, other baseline models exhibit several shortcomings, such as distorted faces or backgrounds, incomplete removal of hands, restoration of incorrect hands (not belonging to the observed baby), and only partial removal of obstructions. This disadvantage is present even for the second-best model \textit{Tuned-runwayml}.

DiffMVR adeptly handles various challenging conditions such as dim lighting and varied object textures and colors, demonstrating its wide applicability in diverse inpainting scenarios. See the Appendix for more details.

\subsection{Ablation Study}
In the inpaint pipeline, we develop two key innovations: the dual-guidance module, which synthesizes fused embeddings from both short-term past and present frames to generate a new combined attention score, and the U-Net module, which designs and integrates a new motion-consistency loss term to guide the denoising process. In this section, we conduct a comprehensive ablation study to assess the effectiveness of having either and both modules in the video object restoring pipeline. 

\subsubsection{Guidance Components Ablation}
\label{subsec:Ablation1}
We contrast the performance of our model with variants that rely solely on a single-image guidance to illustrate the advantages of our multi-frame guidance module. This experiment specifically tests the impact of our innovative approach, which encodes guidance images independently and integrates them through a weighted cross-attention mechanism within the U-Net layers. Since from Table \ref{tab:tabv} segmented masks have better test results in the majority of aspects, we only compare results based on this masking type. As shown in Table \ref{tab:tab3}, excluding either present or prior guidance causes the inpaint result metrics to drop drastically, even worse than baseline models sometimes. Utilizing the current frame as guidance does not enhance the inpainting process, as evidenced by its subpar performance, ranking second to last in comparison to benchmarks in both Table \ref{tab:tab3} and Table \ref{tab:tabv}. By comparing DiffMVR against those restricted to a single type of guidance, we stress the necessity of the dual-image guidance design in our pipeline.

\noindent\subsubsection{Loss Component Ablation}
\label{subsec:Ablation2}

Building upon the findings from Section \ref{subsec:Ablation1}, this ablation study further investigates the cumulative impact of integrating the additional motion loss component into our pipeline. To systematically assess the impact of each component, we conduct experiments under several configurations. Using a single past frame as guidance and using merely denoise loss for training is the baseline setting. We gradually add the designs in: i) baseline + dual-guide, ii) baseline + motion loss, and iii) baseline + dual-guide + motion loss, which is our model, DiffMVR. 

We present the results in Table \ref{tab:ab2}. As expected, adding the motion-consistency loss leads to a lower TC score and higher FVD compared to baseline, even when a single image is used as guidance. Adding motion as a portion of loss enhances temporal smoothness and contributes to a more realistic and accurate video frame restoration, revealed by the scope of changes in the row of \textit{Gap}. Besides, the comparison between \textit{baseline} and DiffMVR shows that incorporating both the motion loss and using dual guidance notably improves the model's performance by $26\%$ on average. This confirms that our approach synergizes efficiently.
\begin{table}[!h]
\centering
\scriptsize
\setlength{\tabcolsep}{2.1pt} 
\renewcommand{\arraystretch}{1.0} 
\begin{tabular}{@{}lllll@{}} 
\toprule
{} & \multicolumn{4}{c}{\textbf{Segmented masks}} \\
\cmidrule(lr){2-5} 
\textbf{Model} & \textbf{FID} & \textbf{SSIM} & \textbf{TC} & \textbf{FVD} \\
\midrule
Dual guide & \textbf{2.10} & \textbf{0.91} & \textbf{0.34} & \textbf{48.05} \\
\cmidrule(lr){1-5} 
\begin{tabular}[c]{@{}l@{}}Single guide \\ (symmetric)\end{tabular} & $2.57 \textcolor{orangecolor}{\scriptstyle \blacktriangle 22.4\%}$ & $0.75 \textcolor{brightgreen}{\scriptstyle \blacktriangledown 17.6\%}$ & $0.42 \textcolor{orangecolor}{\scriptstyle \blacktriangle 23.5\%}$ & $59.51 \textcolor{orangecolor}{\scriptstyle \blacktriangle 23.9\%}$ \\
\begin{tabular}[c]{@{}l@{}}Single guide \\ (past frame)\end{tabular} & $2.23 \textcolor{orangecolor}{\scriptstyle \blacktriangle 6.2\%}$ & $0.77 \textcolor{brightgreen}{\scriptstyle \blacktriangledown 15.4\%}$ & $0.38 \textcolor{orangecolor}{\scriptstyle \blacktriangle 11.8\%}$ & $60.80 \textcolor{orangecolor}{\scriptstyle \blacktriangle 26.5\%}$\\
\begin{tabular}[c]{@{}l@{}}Single guide \\ (present frame)\end{tabular} & $2.69 \textcolor{orangecolor}{\scriptstyle \blacktriangle 28.1\%}$ & $0.74 \textcolor{brightgreen}{\scriptstyle \blacktriangledown 18.7\%}$ & $0.39 \textcolor{orangecolor}{\scriptstyle \blacktriangle 14.7\%}$ & $72.79 \textcolor{orangecolor}{\scriptstyle \blacktriangle 51.5\%}$ \\
\bottomrule
\end{tabular}
\caption{Quantitative ablation test on the Baby dataset, highlighting that the design of multi-guidance achieves the best performance. The motion loss is included throughout this comparison test. The $\textcolor{orangecolor}{\blacktriangle}$/$\textcolor{brightgreen}{\blacktriangledown}$ indicates a relative increase/decrease in metric score compared to Dual guide (\textit{DiffMVR}).}
\label{tab:tab3}
\end{table}

\begin{table}[H]
\centering
\scriptsize
\setlength{\tabcolsep}{1.25pt} 
\renewcommand{\arraystretch}{1.05}
\begin{tabular}{@{}l p{1.35cm} p{1.35cm} p{1.35cm} p{1.35cm}@{}}
\toprule
{} & \multicolumn{4}{c}{\textbf{Segmented Masks}} \\
\cmidrule(lr){2-5} 
\textbf{Configuration} & \textbf{FID} & \textbf{SSIM} & \textbf{TC} & \textbf{FVD} \\
\midrule
baseline & 2.86 & 0.68 & $0.41$ & 65.92 \\
baseline + dual & $2.32 \textcolor{brightgreen}{\scriptstyle \blacktriangledown 18.9\%}$ & $0.74 \textcolor{orangecolor}{\scriptstyle \blacktriangle 8.8\%}$ & $0.38 \textcolor{brightgreen}{\scriptstyle \blacktriangledown 7.3\%}$ & $61.57 \textcolor{brightgreen}{\scriptstyle \blacktriangledown 6.6\%}$ \\
baseline + motion & $2.23 \textcolor{brightgreen}{\scriptstyle \blacktriangledown 22.1\%}$ & $0.77 \textcolor{orangecolor}{\scriptstyle \blacktriangle 13.2\%}$ & $0.38 \textcolor{brightgreen}{\scriptstyle \blacktriangledown 7.3\%}$ & $60.80 \textcolor{brightgreen}{\scriptstyle \blacktriangledown 7.8\%}$ \\
\begin{tabular}[c]{@{}l@{}}DiffMVR: \\ baseline + dual + motion\end{tabular} & $\textbf{2.10} \textcolor{brightgreen}{\scriptstyle \blacktriangledown 26.6\%}$ & $\textbf{0.91} \textcolor{orangecolor}{\scriptstyle \blacktriangle 33.8\%}$ & $\textbf{0.34} \textcolor{brightgreen}{\scriptstyle \blacktriangledown 17.1\%}$ & $\textbf{48.05} \textcolor{brightgreen}{\scriptstyle \blacktriangledown 27.1\%}$ \\
\midrule 
Gap (\%) & 9.48 & 22.97 & 10.52 & 21.96 \\ 
\bottomrule
\end{tabular}
\caption{A pervasive ablation test on the Baby dataset, which exemplifies the impact of gradually adding motion-consistency loss to different guidance configurations. The results highlight the combined effect of our innovations in enhancing video inpainting performance. The $\textcolor{orangecolor}{\blacktriangle}$/$\textcolor{brightgreen}{\blacktriangledown}$ indicates a relative increase/decrease in metric score compared to baseline. \textit{Gap} refers to the extent by which \textit{DiffMVR: baseline + dual + motion} outperforms \textit{baseline + dual}.}
\label{tab:ab2}
\end{table}

\section{Conclusions}
In this study, we introduced DiffMVR, a multi-image guided video inpainting model designed to restore complex details in video sequences of varying lengths, effectively leveraging both short-term and spatial-temporal pixel information.

Our experimental results demonstrate that DiffMVR offers promising performance, surpassing all baseline models in both visual quality and quantitative metrics. This success underscores the model's proficiency in handling intricate video restoration tasks. However, opportunities for further refinement exist. One area for potential enhancement is the optimization of the weighting factor \(\lambda\). Determining the optimal \(\lambda\) that aligns with user preferences and specific application requirements remains a challenge and a promising direction for future research.

Overall, DiffMVR showcases significant robustness and efficacy in video-level media restoration. We believe that this work not only advances the field of video inpainting but also lays a foundation for future enhancements. It is our hope that DiffMVR will catalyze further innovations in video processing technologies and inspire downstream video-level inpainting tasks across various domains.

\section*{Acknowledgments}

This research is funded by the U.S. National Science Foundation (NSF) under Grant TB-Inserted-upon-acceptance. 

{
    \small
    \bibliographystyle{ieee}
    \bibliography{main}
}

\end{document}